\documentclass[letterpaper,10pt,journal,twoside]{IEEEtran}

\usepackage{amsmath,amssymb}
\usepackage{graphicx}
\usepackage{booktabs}
\usepackage{multirow}
\usepackage{float}
\usepackage{siunitx}
\usepackage[table]{xcolor}
\usepackage{cite}
\usepackage{url}
\usepackage{algorithm}
\usepackage[noend]{algpseudocode}
\usepackage{tikz}
\usetikzlibrary{arrows.meta}
\usepackage{pgfplots}
\usepackage{pgfplotstable}
\usepgfplotslibrary{groupplots}
\definecolor{cTIflat}{HTML}{1F4E79}
\definecolor{cTIfac}{HTML}{6A9FD4}
\definecolor{cBone}{HTML}{3D3C39}
\definecolor{cBp}{HTML}{E07A1F}
\definecolor{cBs}{HTML}{8E44AD}
\definecolor{cBzero}{HTML}{9E9D98}
\definecolor{cWP}{HTML}{4C78A8}
\definecolor{cIG}{HTML}{D1495B}
\definecolor{cBL}{HTML}{2A9D8F}
\definecolor{cGrid}{HTML}{E4E3DF}
\pgfplotsset{
  compat=1.18,
  every axis/.append style={
    font=\footnotesize, tick label style={font=\scriptsize}, label style={font=\footnotesize},
    title style={font=\footnotesize, yshift=-3pt}, legend style={font=\scriptsize, draw=none, fill=none},
    grid=major, major grid style={cGrid}, axis line style={black!60}, tick style={black!60},
    every axis title/.style={at={(0,1)}, anchor=south west},
  },
  pTIflat/.style={cTIflat, very thick, mark=*, mark size=1.4pt},
  pTIfac/.style={cTIfac, thick, dashed, mark=o, mark size=1.4pt},
  pBone/.style={cBone, thick, mark=square*, mark size=1.2pt},
  pBp/.style={cBp, thick, mark=triangle*, mark size=1.6pt},
  pBs/.style={cBs, thick, mark=diamond*, mark size=1.6pt},
  pBzero/.style={cBzero, thick, densely dashed, mark=x, mark size=1.8pt},
  pWP/.style={cWP, thick, mark=*, mark size=1.2pt},
  pIG/.style={cIG, thick, mark=square*, mark size=1.1pt},
  pBL/.style={cBL, thick, mark=triangle*, mark size=1.5pt},
  ebar/.style={error bars/.cd, y dir=both, y explicit, error bar style={line width=0.6pt}, error mark=none, /pgfplots/.cd},
}

\input{figures/tabledata}
\usepackage[hidelinks]{hyperref}
\usepackage{orcidlink}

\newcommand{\G}{G}
\newcommand{\Kset}{\mathcal{K}}
\newcommand{\Tgrid}{\mathcal{T}}

\newtheorem{proposition}{Proposition}
\algrenewcommand\algorithmicindent{1.0em}

\hypersetup{pdftitle={Time-Indexed Skill Success Models for Latency-Aware Skill Selection and Timing in Dynamic Manipulation}, pdfauthor={Mohsen Omidi, Maria Chiara Leva, Brian Vaughan}}
\begin{document}

\title{Time-Indexed Skill Success Models for Latency-Aware Skill Selection and Timing in Dynamic Manipulation}

\author{Mohsen~Omidi$^{1,*}$, Maria~Chiara~Leva$^{2}$,
and Brian~Vaughan$^{1}$%
\thanks{This work has been submitted to the IEEE for possible publication. Copyright may be transferred without notice, after which this version may no longer be accessible.}%
\thanks{$^{1}$M. Omidi and B. Vaughan are with the
Virtual Interaction Research Lab (VIRaL), School of Media,
Technological University Dublin, Dublin, Ireland.
E-mails: mohsen.omidi@tudublin.ie,
brian.vaughan@tudublin.ie.
$^{2}$M. C. Leva is with the School of Food Science and
Environmental Health, Technological University Dublin,
Dublin, Ireland.
E-mail: mariachiara.leva@tudublin.ie.
$^{*}$Corresponding author: M. Omidi.}%
}
\markboth{Preprint submitted to IEEE Robotics and Automation Letters}%
{Omidi \MakeLowercase{\textit{et al.}}: Time-Indexed Skill Success Models for Latency-Aware Skill Selection}
\maketitle

\begin{abstract}
Robots that manipulate moving objects act on observations that are outdated when the chosen skill starts, and
the best skill can change within a few hundred milliseconds. Skill competence models, however, predict success for a
launch at the moment of observation. We propose time-indexed skill success
models, which predict whole-task success when a skill is launched after a given delay. Trained on rollouts branched
at a grid of launch delays, they let a receding-horizon planner select both the skill and its launch time under fixed
or random latency. In a MuJoCo study in which a UR5e captures a sliding puck with three skills, the planner exceeded
a static competence model by 2.8 to 6.2 percentage points at 100 to 300\,ms of latency, including when that model
was evaluated on point or sampled forecasts of the launch state. Across five complete runs (three training seeds and
two distribution shifts), all 80 paired 95\% intervals against four baselines excluded zero. On delayed launches,
the time-indexed model was calibrated (expected calibration error 0.009 against 0.239), and pausing the world during
latency overstated success by up to 23 points. On a real UR5e that captures a rolling ball after 0.2\,s of injected
latency, a static model trained in a hardware-matched simulator predicted 99\% success on a prespecified subset of 35
delayed launches that it rated at least 0.8. Of these, 63\% succeeded (95\% interval 46 to 77\%), and the
time-indexed model predicted 56\%. In this single-skill, predictable regime, a matched physics-based forecast was
also well calibrated.
\end{abstract}


\section{Introduction}
\label{sec:intro}

\IEEEPARstart{R}{obots} act on information that is already old. Between the exposure of a camera frame and the
first motion of a skill, the system transfers the image, estimates the state, runs inference and sends the command.
For current vision-language-action policies, this end-to-end latency is about 100 to 300\,ms
\cite{black2025rtc,lu2026faster,wei2026f2f}. When the object is at rest, the delay costs only time, but when
it moves, the robot acts in a world that differs from the one it observed. For example, a puck sliding at
0.5\,m/s travels 5\,cm in 100\,ms, and an interception that was feasible at the moment of observation may no longer
be feasible when the skill starts.

\begin{figure*}[t]
\centering
\begin{minipage}[b]{0.45\textwidth}
\centering
\begin{tikzpicture}[x=5.0cm, y=5.0cm, font=\scriptsize]
  \fill[gray!9] (-0.45,0.26) rectangle (0.35,0.74);
  \draw[line width=1.6pt, black!65] (0.35,0.26) -- (-0.45,0.26) -- (-0.45,0.74) -- (0.35,0.74);
  \draw[cIG, line width=1.1pt, densely dashed] (0.35,0.25) -- (0.35,0.75);
  \node[cIG, rotate=90, anchor=north] at (0.36,0.50) {open edge: puck lost};
  \node[anchor=north west, black!55] at (-0.44,0.735) {low-friction channel, 48\,cm $\times$ 80\,cm};
  \draw[black!45, densely dotted, line width=0.8pt] (-0.30,0.60) -- (0.35,0.385);
  \fill[cBone] (-0.30,0.60) circle (0.021);
  \draw[-{Latex[length=2.0mm]}, cBone, line width=0.9pt] (-0.30,0.60) -- (-0.17,0.557);
  \node[anchor=east, cBone] at (-0.325,0.60) {puck at $t$};
  \node[anchor=north, cBone] at (-0.20,0.555) {$\mathbf{v}$};
  \draw[cIG, line width=1.0pt] (-0.035,0.486) -- (0.005,0.526) (-0.035,0.526) -- (0.005,0.486);
  \fill[cBL] (0.1785,0.470) circle (0.009);
  \fill[cBL] (0.1615,0.419) circle (0.009);
  \draw[cWP, line width=1.0pt] (0.275,0.406) circle (0.018);
  \fill[cTIflat!12] (0,0) circle (0.055);
  \draw[cTIflat, line width=0.8pt] (0,0) circle (0.055);
  \node[cTIflat] at (0,0) {UR5e};
  \draw[cTIflat!65, line width=2.6pt, line cap=round] (0.03,0.05) -- (0.11,0.22) -- (0.05,0.40);
  \fill[cTIflat!65] (0.11,0.22) circle (0.014);
  \draw[cTIflat, line width=1.2pt] (0.030,0.405) -- (0.030,0.440);
  \draw[cTIflat, line width=1.2pt] (0.070,0.405) -- (0.070,0.440);
  \node[cTIflat, anchor=west] at (0.13,0.22) {arm at ready pose};
  \draw[black!65, line width=0.8pt] (-0.47,0.01) rectangle (-0.33,0.15);
  \node[black!65] at (-0.40,0.08) {bin};
  \node[anchor=west, cIG] at (-0.47,-0.10) {$\times$ InterceptGrasp: match the puck's velocity and grasp it in motion};
  \node[anchor=west, cBL] at (-0.47,-0.16) {$\bullet\,\bullet$ Block: stop it with a two-post finger barrier, then pick};
  \node[anchor=west, cWP] at (-0.47,-0.22) {$\circ$ WaitPick: wait until the puck rests, then pick};
\end{tikzpicture}
\end{minipage}%
\hfill
\begin{minipage}[b]{0.52\textwidth}
\centering
\begin{tikzpicture}
\begin{axis}[
  width=\linewidth, height=4.6cm, xmin=0, xmax=0.62, ymin=20, ymax=112, ytick={20,40,60,80,100},
  xlabel={launch delay $\tau$ [s]}, ylabel={whole-task success [\%]},
  title={Slow launches, 0.30 to 0.50\,m/s ($n=579$)},
  legend style={at={(0.02,0.03)}, anchor=south west}, legend cell align=left]
  \addplot[pWP] table[x=tau, y=k0, col sep=comma] {\tblpgforacleslow};
  \addplot[pIG] table[x=tau, y=k1, col sep=comma] {\tblpgforacleslow};
  \addplot[pBL] table[x=tau, y=k2, col sep=comma] {\tblpgforacleslow};
  \legend{WaitPick, InterceptGrasp, Block}
  \draw[black!55, densely dashed] (axis cs:0,20) -- (axis cs:0,112);
  \draw[black!55, densely dashed] (axis cs:0.2,20) -- (axis cs:0.2,112);
  \node[anchor=north west, align=left, font=\scriptsize] at (axis cs:0.006,111) {a static model\\scores $\tau=0$};
  \node[anchor=north west, align=left, font=\scriptsize] at (axis cs:0.206,111) {the skill actually starts\\after $\ell=200$\,ms};
  \draw[cIG, line width=0.9pt] (axis cs:0,74.27) circle (3.2pt);
  \node[cIG, anchor=south west, font=\scriptsize] at (axis cs:0.008,76.0) {best now};
  \draw[cWP, line width=0.9pt] (axis cs:0.2,72.02) circle (3.2pt);
  \node[cWP, anchor=south west, font=\scriptsize] at (axis cs:0.208,74.0) {best at $\tau=0.2$\,s};
\end{axis}
\end{tikzpicture}
\end{minipage}
\caption{Task and timing problem. Left: A UR5e must capture a puck that slides towards an open edge and place it in a
bin. Every 0.1\,s, it chooses one of three skills and an optional extra wait $\delta$, and a decision on the
observation at time $t$ takes effect at $t+\ell+\delta$, where $\ell$ is the latency. Right: For slow launches,
InterceptGrasp is the best skill if launched at once, but WaitPick is the best after 200\,ms (oracle rollouts of the
held-out episodes).}
\label{fig:concept}
\end{figure*}
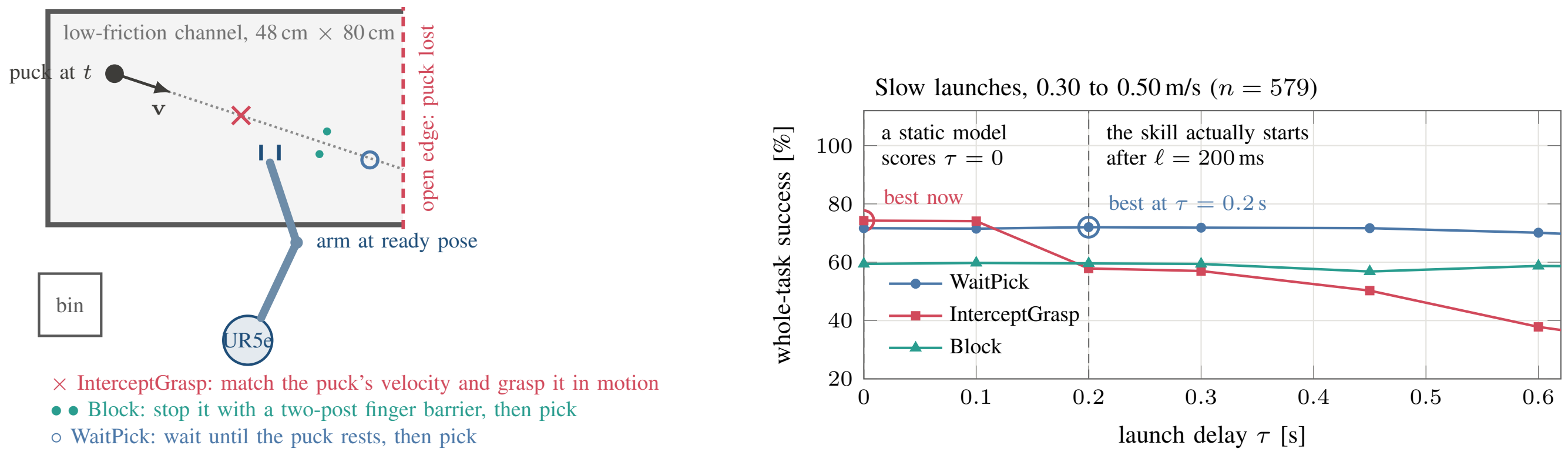

Skill-based and task-and-motion planners select skills with learned competence models, such as initiation
classifiers, affordance values and success discriminators, which map the current state to the probability that a
skill succeeds \cite{konidaris2018skills,bagaria2023effectively,ahn2022can,agia2023stap,liu2024optimistic}. These
models estimate the success of a skill that starts immediately, which is not possible under latency. The planner
needs the probability that the task succeeds if skill $k$ starts after a delay $\tau$, which
includes the unavoidable latency and any extra wait that the planner chooses. Waiting is a real option in dynamic
tasks. A puck that is too fast to grasp may become graspable once friction has slowed it down, while an interception
that is possible now may soon expire (see Fig.~\ref{fig:concept}).

The standard remedy for delay is prediction, as in delay-compensated control \cite{smith1957dead} and in robot
policies that condition on the state rolled forward to execution time \cite{tang2026vlash,liao2025delay}. At the skill
level, the equivalent is to evaluate a static competence model on a forecast of the launch state. We show that this is
exact only in expectation over the forecast distribution and that the error of a point forecast grows with the
forecast spread relative to the tolerance of the skill. With several skills, even sampled forecasts recovered only
part of the loss in our experiments.

In this letter, we therefore learn skill competence as a function of the launch delay. A \emph{time-indexed skill
success model} $\G_k(z,\tau)$ predicts the probability that the whole task succeeds when skill $k$ is launched $\tau$
seconds after the observation of the decision state $z$. The letter makes three contributions. First, we formulate skill
competence as a function of the decision state and the launch delay, and relate it to static models evaluated on
forecasts, which shows when a forecast is sufficient (Section~\ref{sec:problem}). Second, we introduce branched
rollouts that give exact training labels for delayed launches, and a receding-horizon planner that selects the skill
and its launch time under fixed or random latency (Section~\ref{sec:method}). Third, we compare the approach with
static, forecast-based and rule-based planners in a simulation whose clock keeps running during latency, with
replication over seeds and distribution shifts, a calibration analysis and a frozen-clock comparison
(Sections~\ref{sec:setup} and~\ref{sec:results}), and we test the predicted calibration gap on a real UR5e
(Section~\ref{sec:hardware}). Unlike delay-aware continuous policies, the approach works at the level of skill
selection and decides which skill to launch and when from an estimate of whole-task success.

\section{Related Work}
\label{sec:related}

Delay compensation has a long history in control and reinforcement learning. It includes the Smith predictor
\cite{smith1957dead}, the augmentation of the state with pending actions \cite{katsikopoulos2003delays}, and learning under concurrent or random delays
\cite{ramstedt2019realtime,bouteiller2021random}. Recent robot policies handle inference delay within one continuous
policy. Real-time chunking executes the current action chunk while the next one is generated \cite{black2025rtc},
VLASH conditions on the state rolled forward to execution time \cite{tang2026vlash}, and delay-aware diffusion policy
conditions on the measured delay \cite{liao2025delay}. In our setting, no skill runs during the latency and the robot
holds its pose, so the pending-action augmentation reduces to the launch delay, and the pair $(z,\tau)$ becomes the
input of a competence model. Delay also affects evaluation, since offline metrics overstate accuracy when the world
moves during computation \cite{li2020streaming}, and recent dynamic-manipulation benchmarks therefore keep the
simulator running during inference \cite{chen2026reflex}. Planning under deadlines treats time as a budget, for
actions that expire while the search runs \cite{shperberg2019allocating,coles2025dispatch} or for planning effort in a
static world \cite{sung2024deadline}. Here, time is a physical index of how the success of each skill changes with its
start time.

The skill competence models used for planning include learned initiation classifiers \cite{bagaria2023effectively},
symbols learned from the initiation and effect sets of skills \cite{konidaris2018skills}, affordance values
\cite{ahn2022can}, per-skill values composed multiplicatively \cite{agia2023stap}, whole-plan feasibility
classifiers \cite{yang2023piginet} and learned success discriminators \cite{liu2024optimistic}. These models depend on
the state alone, whereas ours adds the launch delay and predicts whole-task success. Confidence calibration of
vision-language-action policies has only recently been studied \cite{zollo2025vlacalibration}. In dynamic
manipulation, dynamic grasping plans reachable grasps of moving objects \cite{akinola2021dynamic}, and projectile
interception has been addressed with belief-space planning \cite{olin2026think} and with planning that offsets a
short delay by forecasting \cite{chen2026becoming,park2026dpp}. Air hockey combines a sliding puck, loop delay and a
choice among several actions, which the leading teams implemented as hand-crafted state machines
\cite{liu2024airhockey}. Skill choice and timing have been studied together in table tennis \cite{wang2017anticipatory},
and monotone optimal stopping learns when to release a single save policy \cite{zhang2026mos}. There, waiting resolves
uncertainty about an opponent's intent, whereas here the forced delay is the robot's own latency and the planner also
chooses among several skills.

\section{Problem Formulation}
\label{sec:problem}

A robot observes a dynamic scene and, at decision times $t$, selects one of $K$ skills $k\in\Kset$. Once launched, a
skill runs in closed loop until it terminates, and fixed follow-up skills then complete the task. Let $z_t$ denote
the decision state estimated from the observations available at time $t$. In our tasks, $z_t$ contains the estimated
position and velocity of the object, quantities derived from them, and the time left until a deadline, which we call
the slack $s$. A decision
taken on $z_t$ cannot take effect before $t+\ell$, where the latency $\ell$ includes perception, computation and
communication delays. The planner may also wait. It selects a skill $k$ and an extra delay $\delta$ from a finite set
$\Delta\ni0$ of candidate waits, and the skill starts at $t+\tau$ with the launch delay $\tau=\ell+\delta$. The robot
holds a ready pose until launch. We define the \emph{time-indexed skill success model} as
\begin{equation}
\G_k(z,\tau)=\Pr\big(\text{success}\,\big|\,z_t=z,\ \text{$k$ starts at $t+\tau$}\big),
\label{eq:G}
\end{equation}
where success means that the whole task, including the follow-up skills, is completed before the deadline. A static
competence model corresponds to $\G_k(z,0)$, which is the success probability of a launch at the moment of
observation. When the latency is known only through its distribution $p(\ell)$, the planner selects
\begin{equation}
(k^\star,\delta^\star)=\operatorname*{arg\,max}_{k\in\Kset,\ \delta\in\Delta}\ \mathbb{E}_{\ell\sim p}
\big[\G_k(z_t,\ell+\delta)\big].
\label{eq:decision}
\end{equation}
If the latency of the decision is known, \eqref{eq:decision} reduces to maximising $\G_k(z_t,\ell+\delta)$.

The following identity links the two kinds of model.
\begin{proposition}\label{prop:tower}
If the robot does not act on the scene before launch and $z_t$ is a sufficient statistic of the observation history
for the future of the task, then for every skill $k$ and delay $\tau\geq0$
\begin{equation}
\G_k(z,\tau)=\mathbb{E}\big[\G_k(z_{t+\tau},0)\,\big|\,z_t=z\big].
\label{eq:tower}
\end{equation}
\end{proposition}
\begin{IEEEproof}
A launch at $t+\tau$ is an immediate launch from the decision state $z_{t+\tau}$. By sufficiency, its success
probability given the history up to $t+\tau$ is $\G_k(z_{t+\tau},0)$. Since the robot idles, the distribution of
$z_{t+\tau}$ given $z_t$ does not depend on the decision, and the tower property of conditional expectation gives
\eqref{eq:tower}.
\end{IEEEproof}

Proposition~\ref{prop:tower} follows directly from the tower property, but it makes the error of each practical
predictor explicit and shows when a forecast is sufficient. The static model $\G_k(z_t,0)$ ignores the delay. A
static model evaluated on a point forecast $\hat z_\tau=\mathbb{E}[z_{t+\tau}\mid z_t]$ moves the expectation inside
the function, and with $\Sigma_\tau=\operatorname{Cov}(z_{t+\tau}\mid z_t)$ its error is, to second order,
\begin{equation}
\G_k(z_t,\tau)-\G_k(\hat z_\tau,0)\approx\tfrac12\operatorname{tr}\!\big(\nabla^2_z\G_k(\hat z_\tau,0)\,
\Sigma_\tau\big).
\label{eq:jensen}
\end{equation}
This error is large where success changes sharply with the state, for example at the edge of a capture tolerance,
and it grows with the forecast spread, and hence with the delay and the disturbance level. Averaging the static model
over sampled forecasts estimates the right-hand side of \eqref{eq:tower}, but only as well as the forecast
distribution is known and the static model, trained on observed states, extrapolates to forecast states. When $z_t$
only summarises the history, \eqref{eq:tower} is itself approximate. A time-indexed model is instead fitted directly
to the left-hand side on exact samples and needs no forecast model at run time.

\section{Method}
\label{sec:method}

\subsection{Branching rollouts and exact training pairs}
\label{sec:data}
For each training episode, we simulate the pre-launch history once with the robot at its ready pose. At each time
$t_j=t_0+\tau_j$ of a launch grid $\Tgrid=\{\tau_0=0,\tau_1,\dots,\tau_M\}$, we store the decision state $z_j$ and a
complete snapshot of the simulator. From every snapshot, we launch every skill, roll it out with its follow-up skills
until the task succeeds or fails, and record the outcome $y_{jk}\in\{0,1\}$. A rollout launched at $t_j$ is then a
labelled sample for every earlier decision, which gives, per episode and skill, the pairs
\begin{equation}
\mathcal{D}_k=\big\{(z_i,\ \tau_j-\tau_i,\ y_{jk})\ :\ 0\leq i\leq j\leq M\big\}
\label{eq:pairs}
\end{equation}
(see Algorithm~\ref{alg:ti}). Because the robot idles at the same pose in the data and in the online planner, these
pairs are exact samples of the left-hand side of \eqref{eq:tower}, and no off-policy correction is needed. Thus,
$M+1$ rollouts yield $(M+1)(M+2)/2$ pairs, which is 45 for our grid of nine delays. All skills share the same
pre-launch history and perception noise, so their outcomes are paired.

\begin{algorithm}[t]
\caption{Branching data generation and receding-horizon launch planning}
\label{alg:ti}
\footnotesize
\begin{algorithmic}[1]
\Statex \textbf{Offline}, for each training episode:
\State simulate with the robot at its ready pose; at $t_j=t_0+\tau_j$, $j=0,\dots,M$, store $z_j$ and a snapshot
$\sigma_j$
\For{$j=0,\dots,M$ \textbf{and} $k\in\Kset$}
  \State restore $\sigma_j$; launch $k$ and its follow-up skills; $y_{jk}\gets$ task success
  \State \textbf{for} $i=0,\dots,j$: add $(z_i,\ \tau_j-\tau_i,\ y_{jk})$ to $\mathcal{D}_k$
\EndFor
\Statex fit $\G_k$ on $\mathcal{D}_k$ by minimising \eqref{eq:loss}
\Statex \textbf{Online}, on each new decision state $z_t$:
\State $S(k,\delta)\gets\frac1Q\sum_{q=1}^{Q}\G_k(z_t,\ell_q+\delta)$ for all $k\in\Kset$, $\delta\in\Delta$
\State $(k^\star,\delta^\star)\gets\arg\max_{k,\delta}S(k,\delta)$
\If{$\delta^\star=0$ \textbf{or} waited $\geq\delta_{\max}$ \textbf{or} object lost}
  \State launch $k^\star$; it starts after the realised latency
\Else
  \State hold the ready pose; decide again after $\max(\Delta t,\ell)$
\EndIf
\end{algorithmic}
\end{algorithm}

\subsection{Flat and factored models}
\label{sec:models}
We compare two forms of the model. TI-flat learns \eqref{eq:G} directly and trains one classifier $\G_k^\theta(z,\tau)$ per skill by minimising the
cross-entropy
\begin{equation}
\mathcal{L}_k(\theta)=-\!\!\sum_{(z,\tau,y)\in\mathcal{D}_k}\!\!\big[y\log\G_k^\theta+(1-y)\log\big(1-\G_k^\theta
\big)\big].
\label{eq:loss}
\end{equation}
TI-fac factors the model along the skill sequence as
\begin{equation}
\G_k(z,\tau)\approx F_k(z,\tau)\;\mathbb{E}_{x'\sim E_k,\,d\sim D_k}\big[V_k(x',\,s-\tau-d)\big],
\label{eq:factored}
\end{equation}
where $F_k$ is the probability that skill $k$ itself succeeds, $E_k(x'\mid z,\tau)$ is a mixture density over the
terminal location of the skill (the rest position of the puck for Block and the gripper holding the puck otherwise),
$D_k(d\mid z,\tau)$ is a log-normal model of its duration, $s$ is the slack, and $V_k$ is the probability that the
follow-up skills complete the task from $x'$ in the remaining time. We fit $E_k$ and $D_k$ on successful executions
only and estimate the expectation with 32 Monte Carlo samples. TI-mean is an ablation that replaces both
distributions with their means. Static counterparts of $F_k$, $E_k$ and $D_k$ have the same architectures but are
trained on the pairs with $\tau=0$ and have no delay input. Since $V_k$ is shared, TI-fac and the static planner B1
(Section~\ref{sec:planners}) differ only in the delay input. All networks are multilayer perceptrons with three hidden layers of 128 SiLU units
(fewer for $D_k$ and $V_k$) on the position, velocity, speed and deceleration estimate of the puck, the slack and, for
the time-indexed models, $\tau$, and every classifier is temperature-scaled on a validation split of 15\% of the
episodes \cite{guo2017calibration}.

\subsection{Receding-horizon planner}
\label{sec:planner}
Every $\Delta t=0.1$\,s, the planner solves \eqref{eq:decision} over the candidate waits
$\Delta=\{0,0.1,\dots,1.2\}$\,s and, as Algorithm~\ref{alg:ti} shows, launches $k^\star$ as soon as $\delta^\star=0$.
A launch of the best skill is forced after $\delta_{\max}=1.3$\,s of waiting or when the tracker loses the object. When
only $p(\ell)$ is known (the \emph{-dist} variants), the expectation in \eqref{eq:decision} is taken over $Q=8$
equal-probability quantiles $\ell_q$, and $Q=1$ when the realised latency is known.

\section{Experimental Setup}
\label{sec:setup}

\subsection{Task, simulation and skills}
\label{sec:task}
A UR5e arm from MuJoCo Menagerie \cite{zakka2022menagerie} with a parallel
gripper stands beside a low-friction channel (48\,cm $\times$ 80\,cm) with side rails. The channel is closed at the
launch end and open at the far end. A puck (diameter 40\,mm, 50\,g) is launched along the channel, and the robot
must capture it before it leaves the open edge and place it in a bin before a deadline. Each episode draws the start
position, the heading ($\pm28^\circ$), the launch speed (0.30 to 1.00\,m/s), the deadline (2.5 to 6.5\,s after
launch), and the sliding friction (0.012 to 0.045) and rail restitution, which the planner never observes. A simulated tracker reports the puck state at 90\,Hz with 25\,ms latency and 2\,mm noise. We simulate the
task with MuJoCo and MuJoCo Warp \cite{todorov2012mujoco,mujocowarp2025} under a clock that never pauses, so the
injected latency and the computation time of each planner (1 to 4\,ms per decision) are charged by advancing the world before
the chosen skill starts.

Three analytic skills run in closed loop on the tracker estimate. WaitPick hovers above the predicted rest point and
grasps the puck once it is slower than 3\,cm/s. This skill is slow but almost insensitive to timing. InterceptGrasp
moves the gripper to the earliest point on the predicted path that it can reach before the puck, matches the velocity
of the puck and grasps it in motion. If no reachable point exists when the skill starts, it fails at once, and we call
such a launch \emph{stale}. Block places the open fingers on the predicted path as a two-post barrier, retreats
compliantly at impact and then picks up the stopped puck. A Place skill finally puts the puck in the bin. We use
analytic skills to separate the planning question from skill learning.

\subsection{Planners}
\label{sec:planners}
Within each model family, the static and time-indexed versions use matched architectures, and the static
versions have no delay input. As references,
Always-$k$ launches one fixed skill at once. B0 is a strong hand-made rule built on the feasibility checks of the
skills, which launches InterceptGrasp if its intercept is feasible, Block if a block point exists, and WaitPick
otherwise. B1 evaluates the static factored model $F_k\,\mathbb{E}[V_k]$ at the observed state and launches the best
skill at once, and B1r adds a myopic wait, holding while the best score rises until it exceeds 0.8 or 1.2\,s have
passed. B1p, the skill-level form of predict-then-act \cite{smith1957dead,tang2026vlash,park2026dpp}, evaluates the
static models on a constant-deceleration forecast of the state at $t+\ell+\delta$. B1s averages the static models over
16 forecasts perturbed by empirical residuals of that predictor. These residuals are measured on the training
episodes and pooled by horizon and speed band, so B1s estimates the right-hand side of \eqref{eq:tower} without
assuming a shape for the forecast error. B1p, B1s and the three time-indexed planners (TI-flat, TI-fac and TI-mean)
all re-plan every 0.1\,s over the same candidate waits (Section~\ref{sec:planner}). Under random latency, B1p-dist
and B1s-dist use the mean or the quantiles of $p(\ell)$.

\subsection{Data and evaluation protocol}
\label{sec:protocol}
The training set has 20{,}000 episodes and 540{,}000 rollouts on the grid
$\Tgrid=\{0,0.1,0.2,0.3,0.45,0.6,0.8,1.0,1.3\}$\,s. Decision times at which the puck had already left the channel or
the tracker had no valid estimate give no pairs, so the set contains 2{,}666{,}595 (decision, launch) pairs instead of
the nominal 2.7 million. All planners are evaluated on the same 2{,}000 held-out episodes
with the same perception noise, so every comparison is paired. We inject fixed latencies
$\ell\in\{0,100,150,200,300\}$\,ms and two random latency models in which every decision draws its own latency. The
first model is uniform between 50 and 300\,ms. The second is heavy-tailed, with $\ell=40+90\,e^{0.55n}$\,ms and
$n\sim\mathcal{N}(0,1)$, capped at 500\,ms (mean 145\,ms, 99th percentile 364\,ms). Rollouts of every skill at every
grid delay of the held-out episodes provide hindsight oracles and the ground truth for calibration. We report
differences in success with 95\% intervals from 2{,}000 paired episode-bootstrap resamples \cite{efron1993bootstrap}.
For predictions $p_i$ of outcomes $y_i$, $i=1,\dots,n$, we report the expected calibration error over 15 equal-width
bins $B_b$ and the Brier score \cite{guo2017calibration,brier1950verification},
\begin{equation}
\mathrm{ECE}=\frac1n\sum_{b}\Big|\sum_{i\in B_b}(y_i-p_i)\Big|,\quad
\mathrm{BS}=\frac1n\sum_{i}(p_i-y_i)^2,
\label{eq:metrics}
\end{equation}
with 95\% intervals from 200 episode-bootstrap resamples. Any temperature refitted for a baseline uses the
even-numbered episodes, and all calibration metrics use the odd-numbered ones. The whole pipeline is repeated in five runs with
training seeds 0, 1 and 2, with faster and wider launches in shift~A (0.40 to 1.20\,m/s, $\pm35^\circ$), and with a
slower and noisier tracker in shift~B (40\,ms latency, 3\,mm noise).

\section{Results}
\label{sec:results}

\subsection{Effect of timing on the best skill}
\label{sec:res_oracle}
In the oracle rollouts of the held-out episodes, which launch exactly at the grid times, the best single skill
launched at the first decision (InterceptGrasp) completed the task in 44.5\% of the episodes, an oracle over skills in
65.6\% and an oracle over skills and launch delays in 71.7\%. The set of skills that would succeed changed with the
launch delay in 71.0\% of the episodes. A launch 0.2\,s later reduced InterceptGrasp to 35.9\% and Block from 43.7\% to
36.7\%, whereas WaitPick did not change (36.8\% and 36.9\%). For slow launches, InterceptGrasp was the best skill if
launched at once and WaitPick after 200\,ms (see Fig.~\ref{fig:concept}), and for fast launches, Block decayed from
34.2\% to 15.7\% within 0.2\,s.

\begin{table}[t]
\centering
\caption{Whole-task success (\%) on the 2{,}000 held-out episodes under fixed latency $\ell$ and under random
per-decision latency known only through its distribution. Best entries in bold. Always-$k$, which launches one fixed
skill at once, reaches at most 44.3\% at 0\,ms (slightly below the oracle rollouts, as every
planned launch is charged at least one 4\,ms control period) and 36.8\% at 300\,ms.}
\label{tab:main}
\footnotesize
\setlength{\tabcolsep}{2.6pt}
\begin{tabular}{@{}lccccccc@{}}
\toprule
 & \multicolumn{5}{c}{Fixed $\ell$ [ms]} & \multicolumn{2}{c}{Random $\ell$} \\
\cmidrule(lr){2-6}\cmidrule(l){7-8}
Planner & 0 & 100 & 150 & 200 & 300 & Heavy & Unif. \\
\midrule
B0 feasibility rule & 63.6 & 55.9 & 48.1 & 46.5 & 41.6 & 50.1 & 48.5 \\
B1 static & 63.6 & 56.9 & 50.3 & 47.5 & 43.3 & 51.7 & 50.1 \\
B1r myopic wait & 61.8 & 54.9 & 47.3 & 45.4 & 40.5 & n/a & n/a \\
B1p point forecast & 63.7 & 55.8 & 49.3 & 46.6 & 43.9 & 50.0 & 48.5 \\
B1s sampled forecasts & 65.0 & 57.0 & 51.9 & 49.1 & 42.8 & 52.1 & 51.1 \\
\midrule
TI-mean (ablation) & 64.1 & 57.8 & 54.0 & 52.1 & 47.5 & n/a & n/a \\
TI-fac (ours) & 65.6 & 58.2 & 55.1 & 52.2 & 47.1 & 55.2 & 53.1 \\
TI-flat (ours) & \textbf{65.8} & \textbf{59.8} & \textbf{55.4} & \textbf{52.8} & \textbf{48.4} & \textbf{55.5} & \textbf{53.9} \\
\bottomrule
\end{tabular}
\end{table}

\begin{figure}[t]
\centering
\begin{tikzpicture}
\begin{axis}[width=8.2cm, height=4.1cm, xmin=-15, xmax=315, xtick={0,100,150,200,300},
  xlabel={injected latency $\ell$ [ms]}, ylabel={TI-flat $-$ baseline [points]}, ymin=-1.5, ymax=11,
  ytick={0,2,4,6,8,10}, legend cell align=left,
  legend style={at={(0.02,0.98)}, anchor=north west, /tikz/every even column/.append style={column sep=5pt}},
  legend columns=4]
  \addplot[pBs, only marks, ebar] table[x expr=\thisrow{latency}*1000-9, y=diff, y error plus=ep, y error minus=em, col sep=comma] {\tblpairedbBs};
  \addplot[pBp, only marks, ebar] table[x expr=\thisrow{latency}*1000-3, y=diff, y error plus=ep, y error minus=em, col sep=comma] {\tblpairedbBp};
  \addplot[pBone, only marks, ebar] table[x expr=\thisrow{latency}*1000+3, y=diff, y error plus=ep, y error minus=em, col sep=comma] {\tblpairedbB};
  \addplot[pBzero, only marks, ebar, solid] table[x expr=\thisrow{latency}*1000+9, y=diff, y error plus=ep, y error minus=em, col sep=comma] {\tblpairedbA};
  \draw[black!60] (axis cs:-15,0) -- (axis cs:315,0);
  \legend{vs B1s, vs B1p, vs B1, vs B0}
\end{axis}
\end{tikzpicture}
\caption{Paired differences in whole-task success between TI-flat and each baseline on the 2{,}000 held-out episodes,
with 95\% bootstrap intervals (markers offset horizontally). TI-flat exceeds every baseline at 100\,ms and above.}
\label{fig:results}
\end{figure}
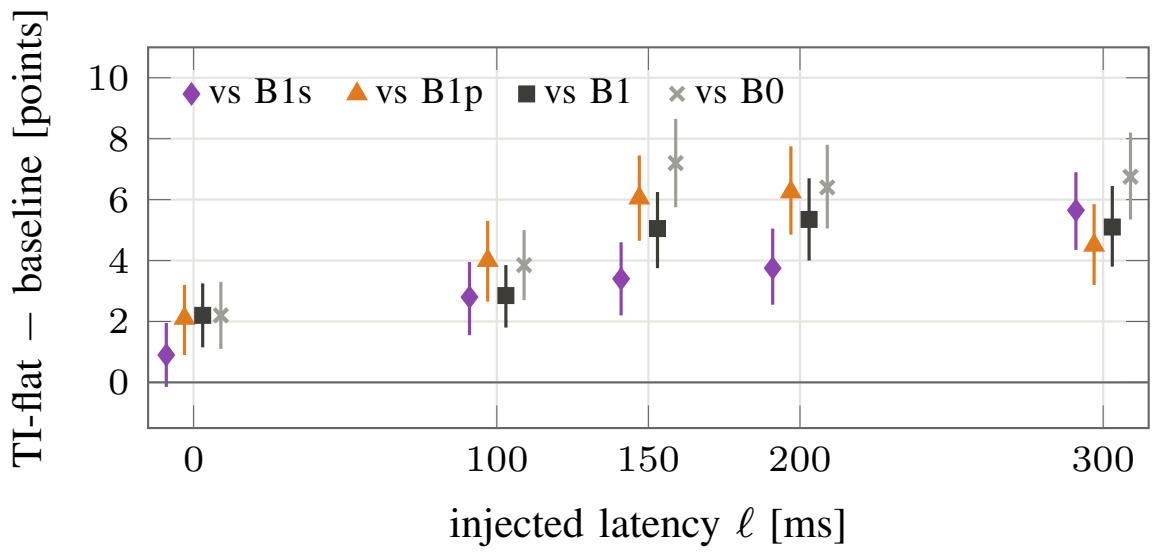

\subsection{Success under fixed and random latency}
\label{sec:res_main}
Table~\ref{tab:main} summarises whole-task success, and Fig.~\ref{fig:results} shows the paired differences. All adaptive planners lost success as
latency increased. However, TI-flat lost the least (from 65.8\% at 0\,ms to 48.4\% at 300\,ms) and was the best
planner at every latency. At 100 to 300\,ms, it exceeded B1 by 2.9 to 5.3 points, B1p by 4.0 to 6.2, B1s by 2.8 to
5.7 and B0 by 3.9 to 7.2, and every 95\% interval was above zero. At 0\,ms, the
advantage was about two points over B1, B1p and B0, and 0.9 points over B1s ($[-0.1,1.9]$). Without latency, a good
static model with sampled forecasts is therefore nearly sufficient, and the remaining gain comes from the option to
wait. TI-fac, which differs from B1 only in the delay input, exceeded B1 by 1.3 to 4.9 points at every latency, which
isolates the contribution of the launch-delay index. The forecasts changed success only slightly, as B1p stayed within
about one point of B1 and B1s gained at most 1.6 points over it.

When every decision drew its own latency and the planners knew only its distribution, TI-flat exceeded B1 by 3.8
points under both latency models, B1s-dist by 3.4 points (heavy-tailed) and 2.8 points (uniform), B1p-dist by 5.5
and 5.4 points, and B0 by 5.4 points, and every paired 95\% interval excluded zero. Knowing the realised latency of each decision
added only 0.9 points $[0.2,1.7]$ and 0.5 points $[-0.4,1.5]$. For the two latency models tested, knowing only the
distribution therefore captured most of the attainable benefit.

\subsection{Source of the gain}
\label{sec:res_mech}
TI-flat waited for at least one decision period in 28 to 36\% of the episodes. At 150\,ms, it succeeded where B1
failed in 140 episodes, and the reverse happened in 39. In 107 of these 140 episodes, B1 had launched InterceptGrasp,
and 102 of these launches were stale. A planner that faces an episode no skill can save may still gamble on
InterceptGrasp. We therefore count \emph{avoidable stale commitments}, which are stale InterceptGrasp launches in
episodes where WaitPick or Block, launched at the same moment, would have completed the task. At 150 to 300\,ms, these
occurred in 5.9 to 6.3\% of the episodes for B1, 7.3 to 7.9\% for B0, 2.7 to 4.0\% for B1p, 1.7 to 2.5\% for B1s and
0.9 to 1.2\% for TI-flat. At 150\,ms, the difference in avoidable stale commitments
between B1 and TI-flat (4.7 points) was close to their success gap (5.1 points). Against B1s, the difference was smaller (1.2 against 3.4
points), which suggests that the time-indexed model also gains by choosing better among the skills that remain
feasible.

\begin{table}[t]
\centering
\caption{ECE of the skill-success predictors on delayed launches ($\tau>0$) of the held-out episodes, with 95\%
episode-bootstrap intervals. The last column gives the mean prediction (\%) on the confident subset, the 3{,}228 InterceptGrasp launches with delays of 150 to 300\,ms that the static model rated at least 0.8.}
\label{tab:calibmain}
\footnotesize
\setlength{\tabcolsep}{2.3pt}
\begin{tabular}{@{}lccc@{}}
\toprule
 & \multicolumn{2}{c}{ECE} & Confident \\
\cmidrule(lr){2-3}
Predictor & InterceptGrasp & Block & subset (\%) \\
\midrule
Time-indexed & 0.009 \scriptsize[0.006, 0.016] & 0.006 \scriptsize[0.005, 0.013] & 72.0 \\
Static & 0.239 \scriptsize[0.226, 0.254] & 0.234 \scriptsize[0.221, 0.249] & 98.8 \\
Static + temperature & 0.119 \scriptsize[0.108, 0.131] & 0.133 \scriptsize[0.119, 0.148] & 66.8 \\
Static, point forecast & 0.126 \scriptsize[0.118, 0.135] & 0.092 \scriptsize[0.083, 0.101] & 71.5 \\
Static, sampled forecasts & 0.059 \scriptsize[0.051, 0.066] & 0.053 \scriptsize[0.045, 0.061] & 70.0 \\
\midrule
Observed success (\%) & & & 71.8 \\
\bottomrule
\end{tabular}
\end{table}

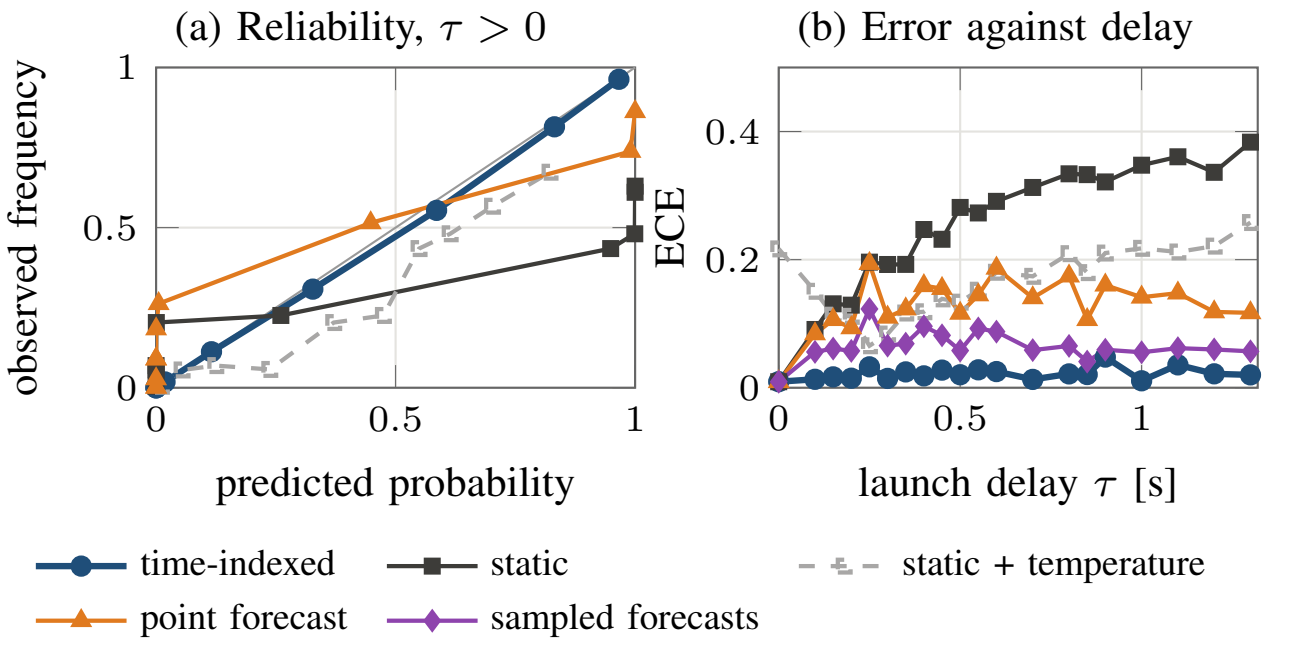
\begin{figure}[t]
\centering
\begin{tikzpicture}
\begin{groupplot}[group style={group size=2 by 1, horizontal sep=0.95cm},
  width=4.75cm, height=3.7cm, legend cell align=left]
  \nextgroupplot[xmin=0, xmax=1, ymin=0, ymax=1, xlabel={predicted probability}, ylabel={observed frequency},
    xtick={0,0.5,1}, ytick={0,0.5,1}, title={(a) Reliability, $\tau>0$},
    legend to name=calmainleg, legend columns=3,
    legend style={/tikz/every even column/.append style={column sep=5pt}}]
  \draw[black!40] (axis cs:0,0) -- (axis cs:1,1);
  \addplot[pTIflat] table[x=pred, y=obs, col sep=comma] {\tblrelInterceptGraspFTI};
  \addplot[pBone] table[x=pred, y=obs, col sep=comma] {\tblrelInterceptGraspFStatic};
  \addplot[cBone!45, thick, dashed, mark=square, mark size=1.2pt] table[x=pred, y=obs, col sep=comma] {\tblrelInterceptGraspFStaticT};
  \addplot[pBp] table[x=pred, y=obs, col sep=comma] {\tblrelInterceptGraspFPred};
  \addlegendimage{pBs}
  \legend{time-indexed, static, static + temperature, point forecast, sampled forecasts}
  \nextgroupplot[xmin=0, xmax=1.32, ymin=0, ymax=0.5, xlabel={launch delay $\tau$ [s]}, ylabel={ECE},
    title={(b) Error against delay}, ylabel shift=-3pt]
  \addplot[pTIflat] table[x=tau, y=TI, col sep=comma] {\tblecebytauInterceptGraspF};
  \addplot[pBone] table[x=tau, y=Static, col sep=comma] {\tblecebytauInterceptGraspF};
  \addplot[cBone!45, thick, dashed, mark=square, mark size=1.2pt] table[x=tau, y=StaticTT, col sep=comma] {\tblecebytauInterceptGraspF};
  \addplot[pBp] table[x=tau, y=Static_predicted, col sep=comma] {\tblecebytauInterceptGraspF};
  \addplot[pBs] table[x=tau, y=Static_sampled, col sep=comma] {\tblecebytauInterceptGraspF};
\end{groupplot}
\end{tikzpicture}

\vspace{1pt}
\ref{calmainleg}
\caption{InterceptGrasp skill-success predictors on held-out launches. (a) Reliability on delayed launches. The
static model is over-confident at both ends, a refitted temperature only flattens it, and the time-indexed model is
close to the diagonal. (b) ECE against the launch delay. Block shows the same pattern.}
\label{fig:calibmain}
\end{figure}

\subsection{Calibration on delayed launches}
\label{sec:res_calib}
On delayed launches (Table~\ref{tab:calibmain} and Fig.~\ref{fig:calibmain}), the time-indexed model was
calibrated, with an ECE of 0.009 for InterceptGrasp and 0.006 for Block (0.008 and 0.006
for whole-task success). The static model was not (0.239 and 0.234), as it
predicted near-certain success or failure and over-predicted InterceptGrasp success by 14.4 points on average. A
temperature refitted on delayed launches halved its ECE, but at zero delay the same temperature raised the ECE from
about 0.01 to between 0.13 and 0.23. No single temperature therefore works for all delays. The forecast-based predictors performed better, but their errors were still six to sixteen times larger than
those of the time-indexed model. The Brier score, which does not depend on binning, confirms this result, with 0.075
for the time-indexed model, 0.113 for the best forecast-based predictor and 0.241 for the static model on
InterceptGrasp (0.059, 0.097 and 0.235 for Block). The hardware test reproduces the confident subset of Table~\ref{tab:calibmain}, the InterceptGrasp launches
with delays of 150 to 300\,ms that the static model rated at least 0.8. On these launches, the static model predicted 98.8\%, 71.8\% succeeded, and the time-indexed model predicted
72.0\%. The
forecast-based predictors were also close on this subset but erred elsewhere, as launches that the point-forecast
predictor rated near-certain ($p\geq0.9$) succeeded in 78.9\% of the cases and those rated near-hopeless ($p<0.1$) in
10.3\%.

\subsection{Evaluation with a frozen clock}
\label{sec:res_frozen}
We also reran the held-out episodes with the world frozen during latency and computation, so that every skill
started from the state on which it had been chosen, as in a simulator that pauses for the planner. This evaluation
overstated the success of the adaptive planners by 6.2 to 8.6 points at 100\,ms
and by 16.9 to 22.8 points at 300\,ms. Under a frozen clock, every adaptive planner
except the myopic B1r scored between 64.0 and 65.9\% at every nominal latency. Such an evaluation hides most of the differences reported above, including the 5.7-point gap between TI-flat and B1s at 300\,ms. As expected, the
policy that always waits for the puck to stop was not affected ($-0.2$ to $+0.2$ points).

\subsection{Replication and ablations}
\label{sec:res_rep}
Across the five runs, TI-flat exceeded B1s by 2.3 to 6.3 points, B1p by 3.0 to 7.1, B1 by 2.1 to 6.9 and B0 by
3.5 to 10.1 at 100 to 300\,ms, and all 80 paired 95\% intervals were above zero, as were all 40 under random latency
with only the distribution known. The ECE of the time-indexed
model on delayed launches stayed between 0.004 and 0.014 in every run, compared with 0.097 to 0.248 for the static
model. TI-flat matched or exceeded TI-fac (intervals above zero in 7 of the 20 run-latency
cells at 100 to 300\,ms, none below zero), and TI-mean changed success by at most 1.5 points, so learning whole-task
success directly was at least as good as composing per-stage models, whose errors compound. The gain over B1 was largest for slow launches (7.9 to 9.0 points at 150 and 200\,ms), where waiting for the puck to slow down is most often the right decision.

\section{Hardware Test}
\label{sec:hardware}

The hardware test checks the predicted calibration failure on delayed launches. It is not meant to show that the
planner outperforms the baselines on hardware, since such a comparison would require an impractical number of trials
(see Section~\ref{sec:discussion}). We therefore used an inclined-plane version of the task with a single capture
skill.

\input{figures/fig_hw}

\subsection{Task, skill and protocol}
\label{sec:hw_task}
A UR5e with a Robotiq Hand-E gripper captures a golf ball (42.7\,mm) that rolls down a board inclined by about
$3^\circ$ (see Fig.~\ref{fig:hw}a). The fingers are fully open (53\,mm), and a bag behind them catches the ball. A
D455f camera tracks the ball at 90 frames/s, and an extended Kalman filter predicts where the ball will cross a catch
line. The filter uses an acceleration field fitted to 72 recorded rolls (RMS error 2.5\,mm),
\begin{equation}
\ddot x=f_0(x)+c\,\tilde y,\qquad \ddot y=f_1(x)+f_2(x)\,\tilde y+f_3(x)\,\tilde y^{2},
\label{eq:field}
\end{equation}
where $x$ points down the slope, $\tilde y=y-0.24$\,m, and $f_0,\dots,f_3$ are piecewise linear in $x$ (25 parameters
in total). The capture skill moves the hand along the line, at lateral position $y_{\mathrm h}$ and speed $v$, to the
predicted crossing $y^\ast$ under the time-optimal law
\begin{equation}
\begin{aligned}
v_{\mathrm d}&=\operatorname{sgn}(e)\min\!\big(v_{\max},\sqrt{2a_{\max}|e|}\big),\\
a&=\operatorname{clip}\big((v_{\mathrm d}-v)/\Delta t_{\mathrm c},\,-a_{\max},\,a_{\max}\big),
\end{aligned}
\label{eq:law}
\end{equation}
with $e=y^\ast-y_{\mathrm h}$, $v_{\max}=0.5$\,m/s and $a_{\max}=2$\,m/s$^2$, and streams the targets at
$1/\Delta t_{\mathrm c}=125$\,Hz. The decision is taken when the predicted time to the line first falls below a
threshold drawn uniformly between 0.45 and 0.85\,s, so
that the decisions cover a range of ball states. The launch is then delayed by $\tau=0.2$\,s in three quarters of
the trials and not delayed in the others, and the hand holds its ready pose until launch. The operator labelled every
outcome.

\subsection{Hardware-matched simulator}
\label{sec:hw_sim}
Training data come from a simulator of this task that is built from the same components. It includes the field
\eqref{eq:field} with Ornstein-Uhlenbeck acceleration perturbations
$w_{n+1}=\alpha\,w_n+\sqrt{1-\alpha^2}\,\Sigma_w^{1/2}\varepsilon_n$, where $\alpha=e^{-\Delta t/T_w}$, $T_w=0.8$\,s,
$\varepsilon_n\sim\mathcal{N}(0,I)$, and the stationary standard deviations are 0.02 and 0.04\,m/s$^2$ along and
across the slope. It also includes the rate, noise and latency of the
tracker, the same filter and control law, and a per-launch camera-to-robot bias
$b\sim\mathcal{N}(0,(1.5\,\text{mm})^2)$. Two properties of the real system were measured on a separate calibration
set of 35 launches, before any test launch, and built into the simulator. First, the arm follows the streamed target
$y_{\mathrm{cmd}}$ as $T\,\dot y_{\mathrm h}(t)+y_{\mathrm h}(t)=y_{\mathrm{cmd}}(t-\theta)$, with a delay
$\theta=35$\,ms and a lag $T=25$\,ms (RMS error 0.84\,mm). Second, the ball meets the fingers 3.9\,cm before the taught line. At this point, the fingers guide the ball into the
bag if the hand is within about 16\,mm of it, which is much wider than the 5\,mm geometric clearance. The largest
error of a capture was 15.6\,mm, and the smallest error of a miss was 16.9\,mm. A simulated launch therefore succeeds
if, at the time $t_{\mathrm c}$ at which the ball reaches the fingers,
\begin{equation}
\big|y_{\mathrm b}(t_{\mathrm c})-y_{\mathrm h}(t_{\mathrm c})-b\big|\leq w_{\mathrm c},\qquad
w_{\mathrm c}\sim\mathcal{N}\big(16,\,2.4^2\big)\ \text{mm}.
\label{eq:capture}
\end{equation}

Each of 5{,}000 episodes is branched at up to 14 decision times and eight launch delays between 0 and 0.4\,s.
Decision times after the ball had crossed the line, or without a valid filter estimate, are skipped, which leaves
449{,}744 exact (decision, delay) pairs of the nominal 560{,}000. Since no follow-up skill is needed, success here is the capture itself. The static model $F(z)$ is a gradient-boosted
classifier of capture success trained on
zero-delay pairs, while $G(z,\tau)$ is trained
on all pairs. Both models use
ten features of the filter state, including the estimates, their standard deviations and the predicted crossing. The models were then frozen. The transfer also required a faster filter and a corrected arm delay, and on the
robot and in the simulator the target is frozen once the ball reaches the hand, so that the hand cannot push a ball
stopped at a finger.

\subsection{Results}
\label{sec:hw_res}
Of the 52 launches recorded after the models were frozen, three were discarded under a rule fixed beforehand (an
interval between control commands above 20\,ms). One ball stopped at the fingers and entered the bag only after the
hand had pushed it, and we counted it as a miss, which left 49 launches. On the 35 delayed launches that the static
model rated at least 0.8, a subset defined before the test, it predicted 0.99 on average, while 22 succeeded (0.63, 95\% Wilson interval 0.46 to 0.77
\cite{wilson1927probable}) and the time-indexed model predicted 0.56 (see Fig.~\ref{fig:hw}c). Before the test, we had
fixed the criterion that the 95\% interval of the observed rate on these launches should exclude the mean prediction
of the static model, and this criterion was met. Over all 49 launches, the Brier score fell from 0.260 to 0.031 (difference 0.230, 95\%
interval 0.110 to 0.360 from 5{,}000 bootstrap resamples of launches), and, as a secondary measure, the area under the ROC curve rose from 0.84 (95\% interval
0.70 to 0.95) to 1.00 (0.99 to 1.00). Without delay, the static model was correct (10 of 10 captured at a mean prediction of 0.99). It failed
only when 0.2\,s of the time budget passed before the hand moved. With the labels as entered and no launch discarded,
its Brier score still exceeded that of the time-indexed model by 0.177 (0.055 to 0.309).

In this task, a static model evaluated on a forecast of the launch state was also well calibrated. We propagated
the mean and covariance of the filter $\tau$ ahead with the fitted field and evaluated $F$ on the result. This is the
strongest forecast available, because the simulator was built from the same model. It predicted 0.60 on the
confident subset and reached a Brier score of 0.020 (AUC 1.00), slightly below that of the time-indexed model
(difference $-0.011$, 95\% interval $-0.021$ to $-0.003$), which was mildly under-confident in the middle of its range.
As an oracle, $F$ evaluated on the state estimated at launch, from frames that are not available when deciding,
predicted 0.61 and reached 0.006 (difference $-0.024$, $-0.056$ to $-0.005$). This is the
regime in which \eqref{eq:jensen} is small, since the forecast spread at 0.2\,s (a few millimetres) is small compared
with the 16\,mm tolerance.

In the hardware-matched simulator, the forecast-based predictor drifted as the delay and
the disturbance increased. Its ECE over ten bins was 0.010 at 0.2\,s and 0.022 at 0.4\,s, compared with 0.004 and
0.003 for the time-indexed model, and 0.039 compared with 0.007 at 0.4\,s with doubled perturbations (see
Fig.~\ref{fig:hw}b). Inside the launch window of the test, however, the two predictors differed in Brier score by less
than 0.007, and more than 2{,}000 paired launches would be needed to resolve this difference. On
hardware, the static model failed because it scored a delayed launch at the observed state.

\section{Discussion and Limitations}
\label{sec:discussion}

Following \eqref{eq:tower} and \eqref{eq:jensen}, a forecast is a good correction when the object is
predictable, the forecast spread is small compared with the tolerance of the skill, and the outcome of the skill
depends on the state of the object alone. The hardware task meets these conditions, since it has one skill, no
contact before the capture, and a forecast model that also generated the training data. The MuJoCo task does not.
Rail bounces and the motion of the robot after launch make the outcome depend on more than a forecast of the puck,
the static models must extrapolate to forecast states, and the choice among skills amplifies miscalibration. These
explanations are consistent with our results, but we have not tested them separately. The time-indexed model does not
rely on these conditions and can also be used when the decision state is a learned representation that cannot be
propagated.

The gains in success are modest in absolute terms (about 3 to 6 points), because timing changes the right decision in
only a minority of episodes. At 150\,ms, for example, TI-flat and B1 differed in outcome in only 179 of the 2{,}000
episodes. On hardware, confirming even the largest band-level gain (slow launches, 9 points at 200\,ms) would require about 380 trials per planner (80\% power,
two-sided $\alpha=0.05$), whereas the much larger calibration gap could be tested.

This work has several limitations. The skills are analytic controllers, and the robot holds its ready pose while it
waits. Learned skills and pre-positioning during the wait are left for future work, although pre-positioning would
break the exactness of the branched pairs. The models interpolate between nine grid delays, and the simulated state
comes from a tracker rather than from images. The planner is not optimal either, since at 0\,ms it is about six points
below the hindsight oracle over skills and delays. The planner comparison comes from simulation only. The hardware
test covers one skill, one delay and 49 test launches on one day, labelled by one operator who also saw an automatic
label, with a simulator calibrated on 35 separate launches. The mild under-confidence of the time-indexed model
suggests that the simulated disturbance level was higher than the behaviour of the board on the test day.

\section{Conclusion}
\label{sec:conclusion}
In this letter, we argued that a planner should estimate the success of a skill launched after the latency.
Time-indexed skill success models provide this estimate and are learned from exact branched rollouts. In simulation,
they were calibrated on delayed launches where static models were severely over-confident, and a receding-horizon
planner that used them exceeded static, forecast-based and rule-based planners at 100 to 300\,ms and under random
latency. On a real robot, a static model trained in a hardware-matched simulator predicted 99\% success on the 35
delayed launches it rated at least 0.8, of which 63\% succeeded, while the time-indexed model predicted 56\%. For this
single skill and predictable object, a matched physics-based forecast was also well calibrated and had a slightly
lower Brier score than the time-indexed model.

\section*{Acknowledgment}
This work was supported by the ARISE project and by the Culture and Society Research Hub of Technological University Dublin.

\bibliographystyle{IEEEtran}
\bibliography{references}

\clearpage
\onecolumn
\section*{Supplementary Material}
This appendix collects the parameters, latency models and additional tables and figures behind the results reported
above. The letter is self-contained, and this material is provided for completeness.

\let\origtable\table \let\endorigtable\endtable
\renewenvironment{table}[1][]{\origtable[H]}{\endorigtable}
\let\origfigure\figure \let\endorigfigure\endfigure
\renewenvironment{figure}[1][]{\origfigure[H]}{\endorigfigure}
\setcounter{section}{0}\setcounter{table}{0}\setcounter{figure}{0}
\renewcommand{\thesection}{S\arabic{section}}\renewcommand{\theHsection}{S\arabic{section}}
\renewcommand{\thetable}{S\arabic{table}}\renewcommand{\theHtable}{S\arabic{table}}
\renewcommand{\thefigure}{S\arabic{figure}}\renewcommand{\theHfigure}{S\arabic{figure}}

\section{Simulation, data and training parameters}
\label{app:params}

\begin{table}[h]
\centering
\caption{Parameters of the simulated task, the data and the models (configuration of training seed 0; shifts A and
B change only the rows marked).}
\label{tab:params}
\small
\begin{tabular}{@{}llp{8.6cm}@{}}
\toprule
Group & Parameter & Value \\
\midrule
Scene & channel & inner width 0.48\,m ($x$ 0.26 to 0.74\,m in the robot frame), launch line $y=0.45$\,m, open exit edge $y=-0.35$\,m, rails 3\,cm high \\
 & puck & cylinder, diameter 40\,mm, height 24\,mm, mass 50\,g \\
 & bin & centre $(0.08,0.40)$\,m, $14\times14$\,cm \\
 & gripper & parallel jaw, tool-centre offset 160\,mm, stroke 50\,mm, V-groove pads, 0.10\,m/s per finger, 60\,N, 20\,ms actuation delay \\
 & physics & MuJoCo / MuJoCo Warp, time step 2\,ms \\
Robot & arm & UR5e (MuJoCo Menagerie), joint speed $\leq180^\circ$/s, tool speed $\leq1$\,m/s, tool acceleration $\leq6$\,m/s$^2$ \\
 & control & 250\,Hz streaming joint control with damped least-squares inverse kinematics \\
Perception & tracker & 90\,Hz, latency 25\,ms (shift B: 40\,ms), position noise 2\,mm (shift B: 3\,mm), velocity noise 0.02\,m/s \\
Episodes & launch & start $x_0\in[0.32,0.68]$\,m, heading $\pm28^\circ$ (shift A: $\pm35^\circ$), speed 0.30 to 1.00\,m/s (shift A: 0.40 to 1.20\,m/s) \\
 & contact & sliding friction 0.012 to 0.045, contact damping ratio 0.25 to 0.9 (unobserved by the planner) \\
 & deadline & 2.5 to 6.5\,s after launch; episode timeout 8\,s \\
 & decisions & first decision 0.12\,s after launch; launch grid $\Tgrid=\{0,0.1,0.2,0.3,0.45,0.6,0.8,1.0,1.3\}$\,s \\
Skills & WaitPick & grasp once the puck is slower than 3\,cm/s; timeout 4\,s \\
 & InterceptGrasp & velocity matching at 5\,m/s$^2$ along the path, close at 6\,mm offset, squeeze width 18\,mm; timeout 3\,s \\
 & Block & two-post barrier, finger gap 16\,mm, compliant retreat absorbing 50\% of the impact speed; timeout 3\,s \\
Data & training & 20{,}000 episodes, 540{,}000 rollouts, 2{,}666{,}595 (decision, launch) pairs, 15\% of episodes for validation \\
 & evaluation & 2{,}000 held-out episodes (separate seed), identical for all planners and all runs of the same configuration \\
Models & networks & MLP, 3 hidden layers of 128 SiLU units ($D_k$: 96, $V_k$: 64); $E_k$: 3-component MDN; inputs position, velocity, speed, deceleration estimate and slack clipped to $[-2,8]$\,s, standardised \\
 & training & AdamW, learning rate $2\times10^{-3}$, weight decay $10^{-5}$, cosine schedule, $\leq60$ epochs, early stopping (patience 8), temperature scaling on validation \\
Planner & timing & decision period 0.1\,s, candidate waits $\{0,0.1,\dots,1.2\}$\,s, forced launch after 1.3\,s or when the puck is lost \\
 & factored scoring & 32 Monte Carlo samples of $E_k$ and $D_k$ \\
 & B1s & 16 forecasts per candidate; residual pools by horizon (0.05\,s bins) and speed band ($<0.5$, 0.5 to 0.75, $\geq0.75$\,m/s), from training episodes only \\
 & compute charge & fixed per planner (mean measured decision time, 1 to 4\,ms), rounded with $\ell$ to the 4\,ms control period \\
\bottomrule
\end{tabular}
\end{table}

\section{Latency distributions}
\label{app:latency}
Table~\ref{tab:latdist} summarises the two random latency models. Every decision of every episode draws its latency
from a table seeded by the episode and the decision index, so all planners see the same sequence of latencies on the
same episode. The \emph{-dist} planners represent the distribution by eight equal-probability quantiles
(mid-quantiles), listed in the last column.

\begin{table}[h]
\centering
\caption{Random latency models (milliseconds).}
\label{tab:latdist}
\small
\begin{tabular}{@{}lcccccl@{}}
\toprule
Model & Mean & Median & 95th pct. & 99th pct. & Max. & Quantiles used by the -dist planners \\
\midrule
Uniform, 50 to 300\,ms & 175 & 175 & 287 & 297 & 300 & 66, 97, 128, 159, 191, 222, 253, 284 \\
Heavy-tailed, $40+90\,e^{0.55n}$, cap 500\,ms & 145 & 130 & 262 & 364 & 500 & 79, 95, 109, 123, 138, 158, 187, 249 \\
\bottomrule
\end{tabular}
\end{table}

\section{Skill success against the launch delay}
\label{app:skillmap}
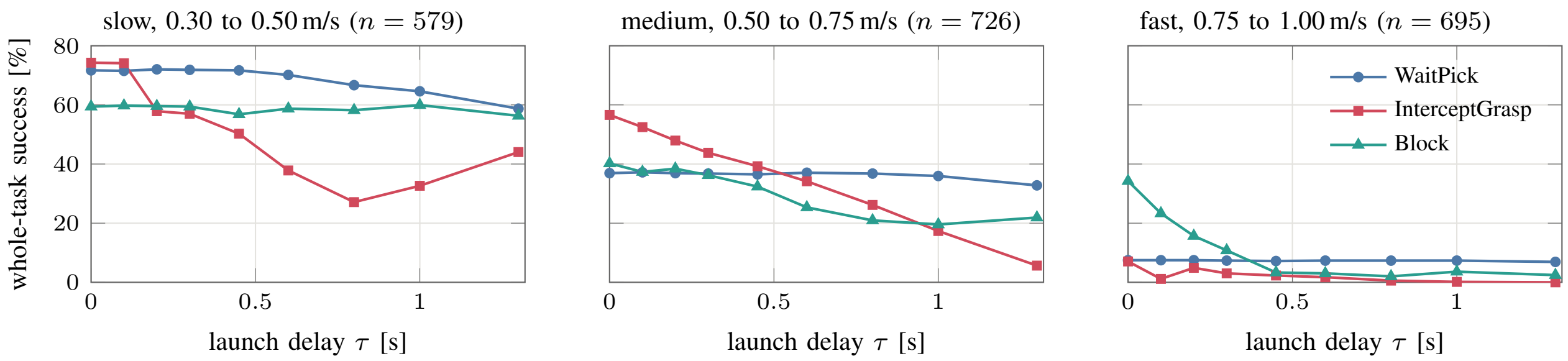
\begin{figure}[t]
\centering
\begin{tikzpicture}
\begin{groupplot}[group style={group size=3 by 1, horizontal sep=0.9cm},
  width=6.2cm, height=4.1cm, xmin=0, xmax=1.32, ymin=0, ymax=80, xlabel={launch delay $\tau$ [s]}]
  \nextgroupplot[ylabel={whole-task success [\%]}, title={slow, 0.30 to 0.50\,m/s ($n=579$)}]
  \addplot[pWP] table[x=tau, y=k0, col sep=comma] {\tblpgforacleslow};
  \addplot[pIG] table[x=tau, y=k1, col sep=comma] {\tblpgforacleslow};
  \addplot[pBL] table[x=tau, y=k2, col sep=comma] {\tblpgforacleslow};
  \nextgroupplot[title={medium, 0.50 to 0.75\,m/s ($n=726$)}, yticklabels={}]
  \addplot[pWP] table[x=tau, y=k0, col sep=comma] {\tblpgforaclemedium};
  \addplot[pIG] table[x=tau, y=k1, col sep=comma] {\tblpgforaclemedium};
  \addplot[pBL] table[x=tau, y=k2, col sep=comma] {\tblpgforaclemedium};
  \nextgroupplot[title={fast, 0.75 to 1.00\,m/s ($n=695$)}, yticklabels={},
    legend style={at={(0.97,0.97)}, anchor=north east}, legend cell align=left]
  \addplot[pWP] table[x=tau, y=k0, col sep=comma] {\tblpgforaclefast};
  \addplot[pIG] table[x=tau, y=k1, col sep=comma] {\tblpgforaclefast};
  \addplot[pBL] table[x=tau, y=k2, col sep=comma] {\tblpgforaclefast};
  \legend{WaitPick, InterceptGrasp, Block}
\end{groupplot}
\end{tikzpicture}
\caption{Whole-task success of each skill against its launch delay, by launch speed (oracle rollouts of the 2{,}000
held-out episodes). The dynamic skills lose value within a few hundred milliseconds, whereas WaitPick barely depends
on timing, so the best skill changes with the delay.}
\label{fig:oracle}
\end{figure}

\begin{figure}[h]
\centering
\includegraphics[width=0.9\linewidth]{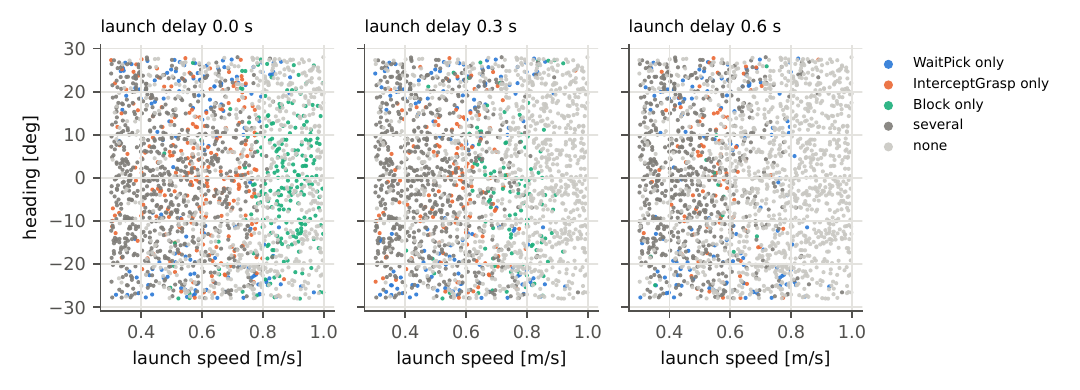}
\caption{Skills that would complete the task in each held-out episode, plotted over launch speed and heading, for
launch delays of 0, 0.3 and 0.6\,s (oracle rollouts). Block-only episodes (fast launches) vanish first; WaitPick-only
episodes persist.}
\label{fig:skillmap}
\end{figure}

Fig.~\ref{fig:oracle} shows the whole-task success of each skill against its launch delay for all three launch-speed
bands. Table~\ref{tab:skillmap} and Fig.~\ref{fig:skillmap} show how the set of skills that would complete the task
changes with the launch delay. The share of episodes in which only Block succeeds collapses from 11.1\% at $\tau=0$ to
0.5\% at 0.6\,s, while the share in which no skill succeeds grows from 34.4\% to 58.2\%.

\begin{table}[t]
\centering
\caption{Share of held-out episodes (\%) by the set of skills that complete the task when launched after delay
$\tau$ (oracle rollouts).}
\label{tab:skillmap}
\setlength{\tabcolsep}{3.5pt}
\begin{tabular}{@{}lccccc@{}}
\toprule
$\tau$ [s] & WaitPick only & InterceptGrasp only & Block only & Several & None \\
\midrule
0.00 & 7.3 & 11.0 & 11.1 & 36.2 & 34.4 \\
0.10 & 7.8 & 9.9 & 8.8 & 33.9 & 39.7 \\
0.20 & 7.9 & 7.8 & 6.8 & 33.5 & 44.0 \\
0.30 & 6.9 & 6.7 & 5.2 & 32.9 & 48.4 \\
0.45 & 7.0 & 6.0 & 2.1 & 31.4 & 53.5 \\
0.60 & 6.2 & 4.1 & 0.5 & 30.9 & 58.2 \\
1.00 & 6.4 & 2.5 & 0.1 & 28.0 & 62.9 \\
\bottomrule
\end{tabular}
\end{table}

\section{Additional results of the main comparison}
\label{app:main}
Fig.~\ref{fig:main} shows the absolute success of the main planners against latency together with the paired
differences, and Table~\ref{tab:paired} lists the paired differences with their intervals. Table~\ref{tab:random}
gives the results under random latency, including the variants told the realised latency of each decision.
Fig.~\ref{fig:mech} shows the avoidable stale commitments of all planners next to the overstatement by a frozen-clock
evaluation. Fig.~\ref{fig:rep} and Table~\ref{tab:rep} summarise the five complete runs.

\begin{figure}[t]
\centering
\begin{tikzpicture}
\begin{groupplot}[group style={group size=2 by 1, horizontal sep=1.5cm},
  width=8.4cm, height=5.0cm, xmin=-15, xmax=315, xtick={0,100,150,200,300}, xlabel={injected latency $\ell$ [ms]}]
  \nextgroupplot[ylabel={whole-task success [\%]}, ymin=38, ymax=78, title={(a) Success under latency},
    legend style={at={(0.99,0.99)}, anchor=north east, /tikz/every even column/.append style={column sep=4pt}},
    legend cell align=left, legend columns=2]
  \addplot[pTIflat] table[x expr=\thisrow{latency}*1000, y=TIflat, col sep=comma] {\tblpgfsuccessvslatency};
  \addplot[pTIfac] table[x expr=\thisrow{latency}*1000, y=TIf, col sep=comma] {\tblpgfsuccessvslatency};
  \addplot[pBs] table[x expr=\thisrow{latency}*1000, y=B1s, col sep=comma] {\tblpgfsuccessvslatency};
  \addplot[pBp] table[x expr=\thisrow{latency}*1000, y=B1p, col sep=comma] {\tblpgfsuccessvslatency};
  \addplot[pBone] table[x expr=\thisrow{latency}*1000, y=B1, col sep=comma] {\tblpgfsuccessvslatency};
  \addplot[pBzero] table[x expr=\thisrow{latency}*1000, y=B0, col sep=comma] {\tblpgfsuccessvslatency};
  \legend{TI-flat (ours), TI-fac (ours), B1s sampled forecasts, B1p point forecast, B1 static, B0 feasibility rule}
  \nextgroupplot[ylabel={TI-flat minus baseline [points]}, ymin=-1.5, ymax=9.5, title={(b) Paired differences, 95\% intervals},
    legend style={at={(0.99,0.02)}, anchor=south east}, legend cell align=left, legend columns=2]
  \addplot[pBs, only marks, ebar] table[x expr=\thisrow{latency}*1000-9, y=diff, y error plus=ep, y error minus=em, col sep=comma] {\tblpairedbBs};
  \addplot[pBp, only marks, ebar] table[x expr=\thisrow{latency}*1000-3, y=diff, y error plus=ep, y error minus=em, col sep=comma] {\tblpairedbBp};
  \addplot[pBone, only marks, ebar] table[x expr=\thisrow{latency}*1000+3, y=diff, y error plus=ep, y error minus=em, col sep=comma] {\tblpairedbB};
  \addplot[pBzero, only marks, ebar, solid] table[x expr=\thisrow{latency}*1000+9, y=diff, y error plus=ep, y error minus=em, col sep=comma] {\tblpairedbA};
  \draw[black!60] (axis cs:-15,0) -- (axis cs:315,0);
  \legend{vs B1s, vs B1p, vs B1, vs B0}
\end{groupplot}
\end{tikzpicture}
\caption{Main comparison on the 2{,}000 held-out episodes. (a) All planners lose success as latency grows, and the
time-indexed planners lose the least. (b) Paired differences between TI-flat and each baseline. TI-flat exceeds every
baseline at 100\,ms and above, including the static model averaged over sampled forecasts (B1s). Markers are offset
horizontally for legibility.}
\label{fig:main}
\end{figure}
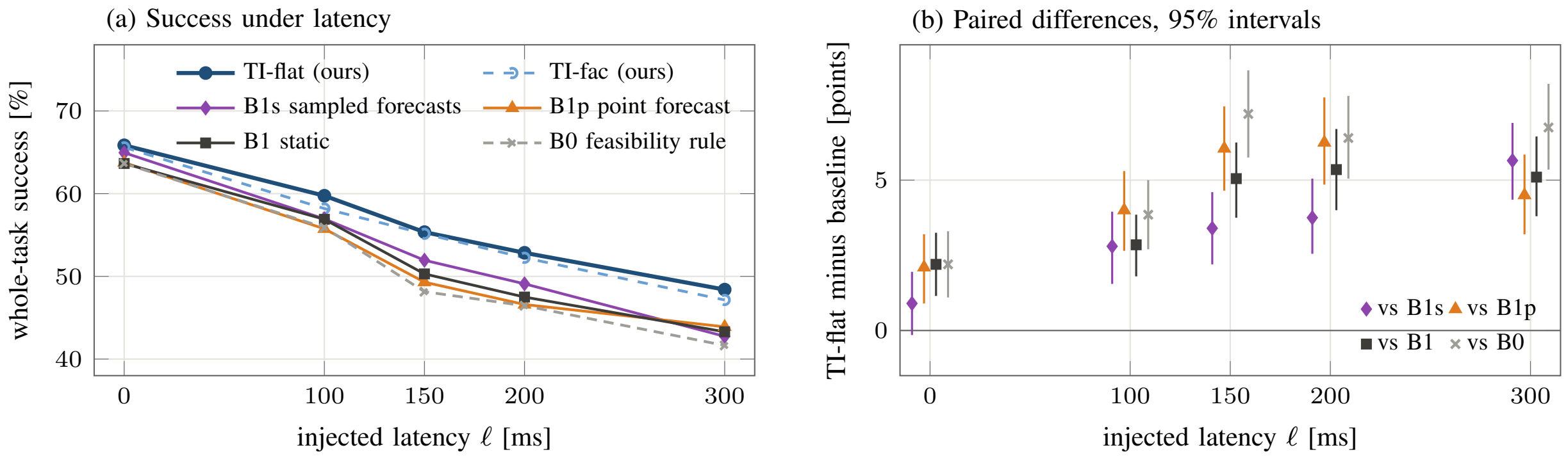
\begin{table}[t]
\centering
\caption{Paired differences in whole-task success (percentage points) with 95\% episode-bootstrap intervals. Bold
entries have intervals that exclude zero. The last row compares models with the same factored structure, which
isolates the effect of the launch-delay index.}
\label{tab:paired}
\setlength{\tabcolsep}{5pt}
\begin{tabular}{@{}lccccc@{}}
\toprule
 & \multicolumn{5}{c}{Injected latency $\ell$ [ms]} \\
\cmidrule(l){2-6}
Comparison & 0 & 100 & 150 & 200 & 300 \\
\midrule
TI-flat $-$ B0 feasibility rule & \textbf{$+$2.2} \scriptsize[1.1, 3.3] & \textbf{$+$3.9} \scriptsize[2.7, 5.0] & \textbf{$+$7.2} \scriptsize[5.8, 8.6] & \textbf{$+$6.4} \scriptsize[5.1, 7.8] & \textbf{$+$6.8} \scriptsize[5.3, 8.2] \\
TI-flat $-$ B1 static model & \textbf{$+$2.2} \scriptsize[1.1, 3.2] & \textbf{$+$2.9} \scriptsize[1.8, 3.9] & \textbf{$+$5.1} \scriptsize[3.7, 6.2] & \textbf{$+$5.3} \scriptsize[4.0, 6.7] & \textbf{$+$5.1} \scriptsize[3.8, 6.5] \\
TI-flat $-$ B1p static, point forecast & \textbf{$+$2.1} \scriptsize[0.9, 3.2] & \textbf{$+$4.0} \scriptsize[2.6, 5.3] & \textbf{$+$6.0} \scriptsize[4.7, 7.4] & \textbf{$+$6.2} \scriptsize[4.9, 7.8] & \textbf{$+$4.5} \scriptsize[3.2, 5.9] \\
TI-flat $-$ B1s static, sampled forecasts & $+$0.9 \scriptsize[$-$0.1, 1.9] & \textbf{$+$2.8} \scriptsize[1.6, 4.0] & \textbf{$+$3.4} \scriptsize[2.2, 4.6] & \textbf{$+$3.8} \scriptsize[2.5, 5.1] & \textbf{$+$5.7} \scriptsize[4.3, 6.9] \\
TI-flat $-$ TI-fac & $+$0.2 \scriptsize[$-$0.8, 1.3] & \textbf{$+$1.6} \scriptsize[0.4, 2.5] & $+$0.2 \scriptsize[$-$0.9, 1.2] & $+$0.6 \scriptsize[$-$0.6, 1.8] & \textbf{$+$1.2} \scriptsize[0.2, 2.2] \\
\midrule
TI-fac $-$ B1 static model & \textbf{$+$1.9} \scriptsize[0.8, 3.0] & \textbf{$+$1.3} \scriptsize[0.1, 2.5] & \textbf{$+$4.9} \scriptsize[3.5, 6.2] & \textbf{$+$4.8} \scriptsize[3.4, 6.1] & \textbf{$+$3.9} \scriptsize[2.5, 5.2] \\
\bottomrule
\end{tabular}
\end{table}

\begin{table}[t]
\centering
\caption{Random per-decision latency. Top: whole-task success (\%). Planners are told either the realised latency of
each decision or only its distribution. Bottom: paired differences (points) with 95\% intervals. Bold intervals
exclude zero.}
\label{tab:random}
\setlength{\tabcolsep}{3pt}
\begin{tabular}{@{}llcc@{}}
\toprule
Planner & Latency & Heavy-tailed & Uniform \\
\midrule
B0 feasibility rule & -- & 50.1 & 48.5 \\
B1 static & -- & 51.7 & 50.1 \\
B1p point forecast & realised & 50.2 & 49.3 \\
B1p point forecast & distribution & 50.0 & 48.5 \\
B1s sampled forecasts & realised & 52.2 & 51.1 \\
B1s sampled forecasts & distribution & 52.1 & 51.1 \\
TI-fac (ours) & realised & 54.9 & 53.8 \\
TI-fac (ours) & distribution & 55.2 & 53.1 \\
TI-flat (ours) & realised & 56.5 & 54.5 \\
TI-flat (ours) & distribution & 55.5 & 53.9 \\
\midrule
\multicolumn{2}{@{}l}{TI-flat $-$ B0 (distribution only)} & \textbf{$+$5.4} \scriptsize[4.1, 6.8] & \textbf{$+$5.4} \scriptsize[4.2, 6.7] \\
\multicolumn{2}{@{}l}{TI-flat $-$ B1 (distribution only)} & \textbf{$+$3.8} \scriptsize[2.5, 5.0] & \textbf{$+$3.8} \scriptsize[2.5, 5.0] \\
\multicolumn{2}{@{}l}{TI-flat $-$ B1p (distribution only)} & \textbf{$+$5.5} \scriptsize[4.0, 7.0] & \textbf{$+$5.4} \scriptsize[4.0, 6.9] \\
\multicolumn{2}{@{}l}{TI-flat $-$ B1s (distribution only)} & \textbf{$+$3.4} \scriptsize[2.1, 4.7] & \textbf{$+$2.8} \scriptsize[1.6, 4.0] \\
\multicolumn{2}{@{}l}{Value of knowing the realised latency} & \textbf{$+$0.9} \scriptsize[0.2, 1.7] & $+$0.5 \scriptsize[$-$0.4, 1.5] \\
\bottomrule
\end{tabular}
\end{table}

\begin{figure}[t]
\centering
\begin{tikzpicture}
\begin{groupplot}[group style={group size=2 by 1, horizontal sep=1.5cm},
  width=8.4cm, height=4.6cm, xmin=-15, xmax=315, xtick={0,100,150,200,300}, xlabel={injected latency $\ell$ [ms]},
  legend cell align=left]
  \nextgroupplot[ylabel={episodes [\%]}, ymin=-0.3, ymax=8.5, title={(a) Avoidable stale commitments}]
  \addplot[pBzero] table[x expr=\thisrow{latency}*1000, y=B0, col sep=comma] {\tblpgfavoidablestalerate};
  \addplot[pBone] table[x expr=\thisrow{latency}*1000, y=B1, col sep=comma] {\tblpgfavoidablestalerate};
  \addplot[pBp] table[x expr=\thisrow{latency}*1000, y=B1p, col sep=comma] {\tblpgfavoidablestalerate};
  \addplot[pBs] table[x expr=\thisrow{latency}*1000, y=B1s, col sep=comma] {\tblpgfavoidablestalerate};
  \addplot[pTIfac] table[x expr=\thisrow{latency}*1000, y=TIf, col sep=comma] {\tblpgfavoidablestalerate};
  \addplot[pTIflat] table[x expr=\thisrow{latency}*1000, y=TIflat, col sep=comma] {\tblpgfavoidablestalerate};
  \nextgroupplot[ylabel={frozen minus real time [points]}, ymin=-1, ymax=25, title={(b) Frozen-clock overstatement},
    legend to name=mechleg, legend columns=7, legend style={/tikz/every even column/.append style={column sep=6pt}}]
  \addplot[pBzero] table[x expr=\thisrow{latency}*1000, y=B0, col sep=comma] {\tblpgffrozenoverstatement};
  \addplot[pBone] table[x expr=\thisrow{latency}*1000, y=B1, col sep=comma] {\tblpgffrozenoverstatement};
  \addplot[pBp] table[x expr=\thisrow{latency}*1000, y=B1p, col sep=comma] {\tblpgffrozenoverstatement};
  \addplot[pBs] table[x expr=\thisrow{latency}*1000, y=B1s, col sep=comma] {\tblpgffrozenoverstatement};
  \addplot[pTIfac] table[x expr=\thisrow{latency}*1000, y=TIf, col sep=comma] {\tblpgffrozenoverstatement};
  \addplot[pTIflat] table[x expr=\thisrow{latency}*1000, y=TIflat, col sep=comma] {\tblpgffrozenoverstatement};
  \addplot[black!35, thick, dotted, mark=none] table[x expr=\thisrow{latency}*1000, y=A0, col sep=comma] {\tblpgffrozenoverstatement};
  \legend{B0, B1, B1p, B1s, TI-fac, TI-flat, always WaitPick}
\end{groupplot}
\end{tikzpicture}

\vspace{2pt}
\ref{mechleg}
\caption{Mechanism and evaluation protocol. (a) Share of episodes in which a planner launched InterceptGrasp, the
skill found no feasible intercept when it started, and WaitPick or Block launched at the same delay would have
completed the task. (b) Success with the world paused during latency and computation minus success with the clock
running, on identical episodes.}
\label{fig:mech}
\end{figure}
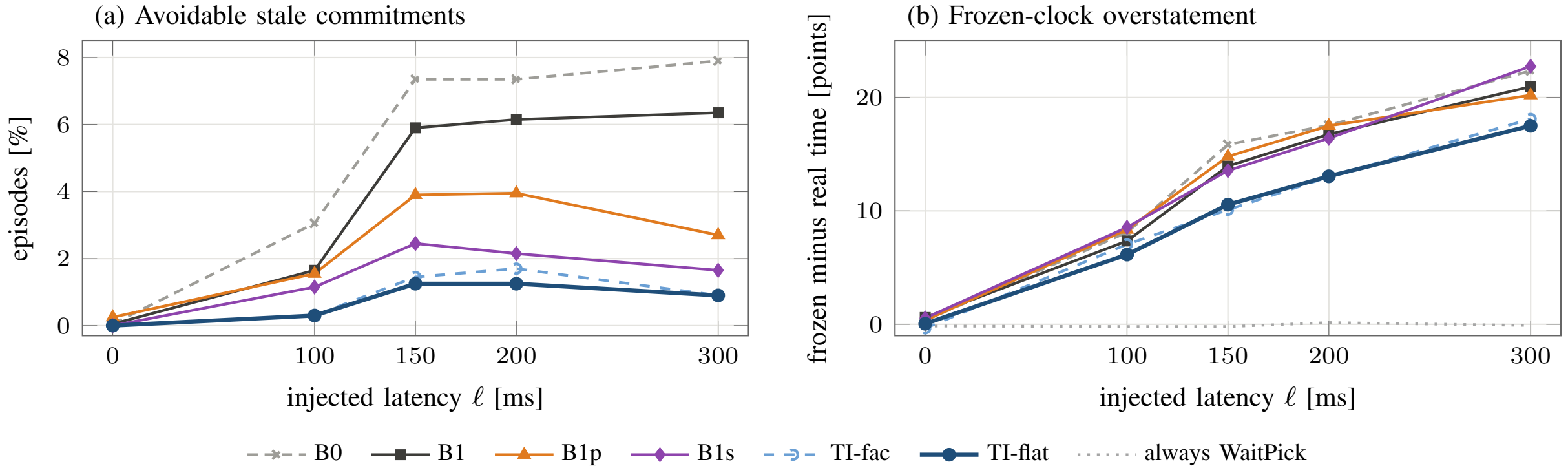
\begin{figure}[t]
\centering
\begin{tikzpicture}
\begin{groupplot}[group style={group size=2 by 1, horizontal sep=1.5cm},
  width=8.4cm, height=4.6cm, xmin=-20, xmax=320, xtick={0,100,150,200,300}, xlabel={injected latency $\ell$ [ms]},
  ymin=-1, ymax=11]
  \nextgroupplot[ylabel={TI-flat minus baseline [points]}, title={(a) vs B1s, sampled forecasts},
    legend style={at={(0.02,0.98)}, anchor=north west}, legend columns=3, legend cell align=left]
  \addplot[cTIflat, only marks, mark=*, mark size=1.5pt, ebar] table[x expr=\thisrow{latency}*1000-10, y=diff, y error plus=ep, y error minus=em, col sep=comma] {\tblrepbBsseedA};
  \addplot[cTIfac, only marks, mark=square*, mark size=1.3pt, ebar] table[x expr=\thisrow{latency}*1000-5, y=diff, y error plus=ep, y error minus=em, col sep=comma] {\tblrepbBsseedB};
  \addplot[cBzero, only marks, mark=triangle*, mark size=1.6pt, ebar] table[x expr=\thisrow{latency}*1000, y=diff, y error plus=ep, y error minus=em, col sep=comma] {\tblrepbBsseedC};
  \addplot[cBp, only marks, mark=diamond*, mark size=1.6pt, ebar] table[x expr=\thisrow{latency}*1000+5, y=diff, y error plus=ep, y error minus=em, col sep=comma] {\tblrepbBsshiftA};
  \addplot[cBs, only marks, mark=pentagon*, mark size=1.6pt, ebar] table[x expr=\thisrow{latency}*1000+10, y=diff, y error plus=ep, y error minus=em, col sep=comma] {\tblrepbBsshiftB};
  \draw[black!60] (axis cs:-20,0) -- (axis cs:320,0);
  \legend{seed 0, seed 1, seed 2, shift A, shift B}
  \nextgroupplot[title={(b) vs B1, static model}]
  \addplot[cTIflat, only marks, mark=*, mark size=1.5pt, ebar] table[x expr=\thisrow{latency}*1000-10, y=diff, y error plus=ep, y error minus=em, col sep=comma] {\tblrepbBseedA};
  \addplot[cTIfac, only marks, mark=square*, mark size=1.3pt, ebar] table[x expr=\thisrow{latency}*1000-5, y=diff, y error plus=ep, y error minus=em, col sep=comma] {\tblrepbBseedB};
  \addplot[cBzero, only marks, mark=triangle*, mark size=1.6pt, ebar] table[x expr=\thisrow{latency}*1000, y=diff, y error plus=ep, y error minus=em, col sep=comma] {\tblrepbBseedC};
  \addplot[cBp, only marks, mark=diamond*, mark size=1.6pt, ebar] table[x expr=\thisrow{latency}*1000+5, y=diff, y error plus=ep, y error minus=em, col sep=comma] {\tblrepbBshiftA};
  \addplot[cBs, only marks, mark=pentagon*, mark size=1.6pt, ebar] table[x expr=\thisrow{latency}*1000+10, y=diff, y error plus=ep, y error minus=em, col sep=comma] {\tblrepbBshiftB};
  \draw[black!60] (axis cs:-20,0) -- (axis cs:320,0);
\end{groupplot}
\end{tikzpicture}
\caption{Replication over five complete runs, each with its own training data, models and evaluation (training seeds
0, 1 and 2, shift A with faster and wider launches, and shift B with slower and noisier perception). Every comparison
at 100\,ms and above has its 95\% interval above zero.}
\label{fig:rep}
\end{figure}
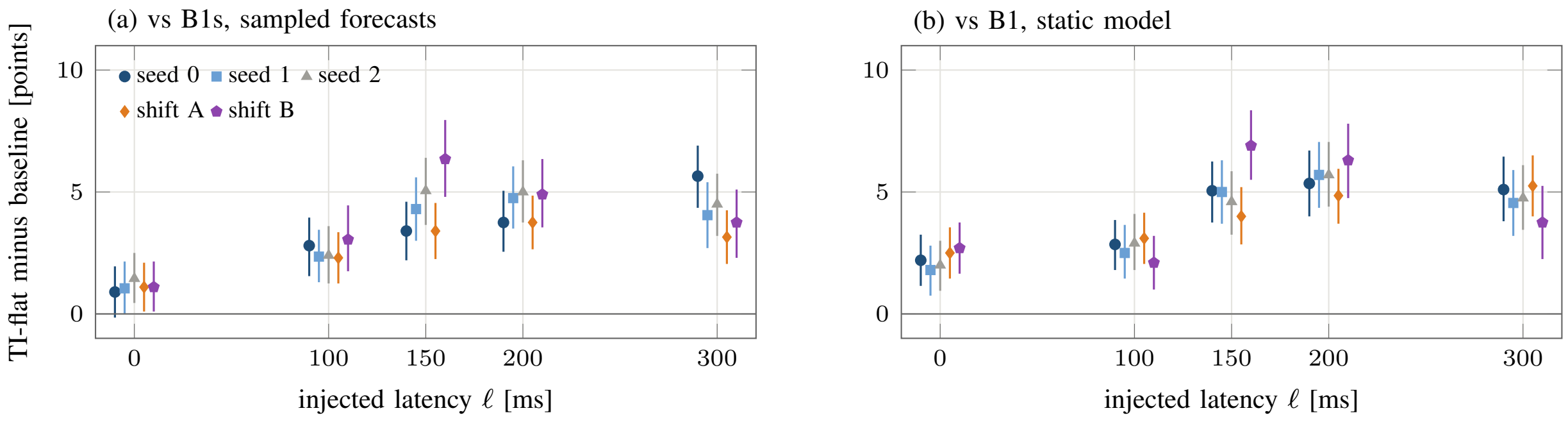
\begin{table}[t]
\centering
\caption{Replication over five complete runs (training seeds 0, 1, 2; parameter shifts A and B). Cells are runs
$\times$ latencies (fixed) or runs $\times$ latency models (random). In the random rows the planners know only the
latency distribution; the last row compares TI-flat told the realised latency with TI-flat told only its
distribution. Positive means that the 95\% interval is above zero.}
\label{tab:rep}
\setlength{\tabcolsep}{2.5pt}
\begin{tabular}{@{}llcc@{}}
\toprule
Comparison & Condition & Positive & Range [pts] \\
\midrule
TI-flat $-$ B1s & fixed, 100 to 300\,ms & 20/20 & $+$2.3 to $+$6.3 \\
TI-flat $-$ B1s & fixed, 0\,ms & 3/5 & $+$0.9 to $+$1.5 \\
TI-flat $-$ B1p & fixed, 100 to 300\,ms & 20/20 & $+$3.0 to $+$7.1 \\
TI-flat $-$ B1p & fixed, 0\,ms & 5/5 & $+$1.9 to $+$3.9 \\
TI-flat $-$ B1 & fixed, 100 to 300\,ms & 20/20 & $+$2.1 to $+$6.9 \\
TI-flat $-$ B1 & fixed, 0\,ms & 5/5 & $+$1.8 to $+$2.7 \\
TI-flat $-$ B0 & fixed, 100 to 300\,ms & 20/20 & $+$3.5 to $+$10.1 \\
TI-flat $-$ B0 & fixed, 0\,ms & 5/5 & $+$2.0 to $+$2.7 \\
TI-flat $-$ TI-fac & fixed, 100 to 300\,ms & 7/20 & $-$0.4 to $+$1.9 \\
TI-flat $-$ B1s & random, distribution & 10/10 & $+$1.6 to $+$3.5 \\
TI-flat $-$ B1 & random, distribution & 10/10 & $+$3.0 to $+$4.3 \\
TI-flat $-$ B1p & random, distribution & 10/10 & $+$2.4 to $+$6.1 \\
TI-flat $-$ B0 & random, distribution & 10/10 & $+$3.8 to $+$6.2 \\
Knowing the realised latency & random, both & 1/10 & $+$0.1 to $+$0.9 \\
\bottomrule
\end{tabular}
\end{table}

\section{Calibration}
\label{app:calib}
Table~\ref{tab:calib} gives the expected calibration error of skill-success and whole-task predictors for both
dynamic skills, and the mean predictions on the launches that the static model rates at least 0.8.
Fig.~\ref{fig:calib} shows the reliability diagrams and the error against the launch delay for both skills, and
Table~\ref{tab:calibfull} the full metrics.

\begin{table}[t]
\centering
\caption{Calibration on delayed launches ($\tau>0$) of held-out episodes. Top: expected calibration error (15
equal-width bins) with 95\% episode-bootstrap intervals, for skill success (F) and whole-task success (G). Bottom: mean
predictions of each model against the observed rate (\%) on launches 150 to 300\,ms after the observation on which
the static model predicts at least 80\% skill or task success.}
\label{tab:calib}
\setlength{\tabcolsep}{5pt}
\begin{tabular}{@{}lcccc@{}}
\toprule
Predictor & InterceptGrasp, F & InterceptGrasp, G & Block, F & Block, G \\
\midrule
Time-indexed & 0.009 \scriptsize[0.006, 0.016] & 0.008 \scriptsize[0.006, 0.014] & 0.006 \scriptsize[0.005, 0.013] & 0.006 \scriptsize[0.005, 0.012] \\
Static & 0.239 \scriptsize[0.226, 0.254] & 0.197 \scriptsize[0.184, 0.211] & 0.234 \scriptsize[0.221, 0.249] & 0.116 \scriptsize[0.105, 0.130] \\
Static + temperature & 0.119 \scriptsize[0.108, 0.131] & 0.142 \scriptsize[0.130, 0.152] & 0.133 \scriptsize[0.119, 0.148] & 0.084 \scriptsize[0.076, 0.097] \\
Static, point forecast & 0.126 \scriptsize[0.118, 0.135] & 0.082 \scriptsize[0.074, 0.089] & 0.092 \scriptsize[0.083, 0.101] & 0.057 \scriptsize[0.051, 0.065] \\
Static, sampled forecasts & 0.059 \scriptsize[0.051, 0.066] & 0.051 \scriptsize[0.044, 0.059] & 0.053 \scriptsize[0.045, 0.061] & 0.028 \scriptsize[0.023, 0.035] \\
\bottomrule
\end{tabular}

\vspace{4pt}
\begin{tabular}{@{}lcccccc@{}}
\toprule
Static model confident ($p\geq0.8$) & Launches & Static & Observed & Time-indexed & Point forecast & Sampled forecasts \\
\midrule
InterceptGrasp, skill & 3,228 & 98.8 & 71.8 & 72.0 & 71.5 & 70.0 \\
InterceptGrasp, task & 2,370 & 97.5 & 69.3 & 70.3 & 70.3 & 68.0 \\
Block, skill & 5,001 & 99.4 & 85.1 & 85.2 & 85.1 & 83.4 \\
Block, task & 2,187 & 93.1 & 80.7 & 81.1 & 81.8 & 78.7 \\
\bottomrule
\end{tabular}
\end{table}

\begin{figure}[t]
\centering
\begin{tikzpicture}
\begin{groupplot}[group style={group size=4 by 1, horizontal sep=1.05cm},
  width=4.75cm, height=4.6cm, legend cell align=left]
  \nextgroupplot[xmin=0, xmax=1, ymin=0, ymax=1, xlabel={predicted probability}, ylabel={observed frequency},
    title={(a) InterceptGrasp, $\tau>0$}, legend to name=calleg, legend columns=5,
    legend style={/tikz/every even column/.append style={column sep=6pt}}]
  \draw[black!40] (axis cs:0,0) -- (axis cs:1,1);
  \addplot[pTIflat] table[x=pred, y=obs, col sep=comma] {\tblrelInterceptGraspFTI};
  \addplot[pBone] table[x=pred, y=obs, col sep=comma] {\tblrelInterceptGraspFStatic};
  \addplot[cBone!45, thick, dashed, mark=square, mark size=1.2pt] table[x=pred, y=obs, col sep=comma] {\tblrelInterceptGraspFStaticT};
  \addplot[pBp] table[x=pred, y=obs, col sep=comma] {\tblrelInterceptGraspFPred};
  \addlegendimage{pBs}
  \legend{time-indexed, static, static + temperature, static on point forecast, static on sampled forecasts}
  \nextgroupplot[xmin=0, xmax=1, ymin=0, ymax=1, xlabel={predicted probability}, title={(b) Block, $\tau>0$}]
  \draw[black!40] (axis cs:0,0) -- (axis cs:1,1);
  \addplot[pTIflat] table[x=pred, y=obs, col sep=comma] {\tblrelBlockFTI};
  \addplot[pBone] table[x=pred, y=obs, col sep=comma] {\tblrelBlockFStatic};
  \addplot[cBone!45, thick, dashed, mark=square, mark size=1.2pt] table[x=pred, y=obs, col sep=comma] {\tblrelBlockFStaticT};
  \addplot[pBp] table[x=pred, y=obs, col sep=comma] {\tblrelBlockFPred};
  \nextgroupplot[xmin=0, xmax=1.32, ymin=0, ymax=0.5, xlabel={launch delay $\tau$ [s]}, ylabel={ECE},
    title={(c) InterceptGrasp}, ylabel shift=-3pt]
  \addplot[pTIflat] table[x=tau, y=TI, col sep=comma] {\tblecebytauInterceptGraspF};
  \addplot[pBone] table[x=tau, y=Static, col sep=comma] {\tblecebytauInterceptGraspF};
  \addplot[cBone!45, thick, dashed, mark=square, mark size=1.2pt] table[x=tau, y=StaticTT, col sep=comma] {\tblecebytauInterceptGraspF};
  \addplot[pBp] table[x=tau, y=Static_predicted, col sep=comma] {\tblecebytauInterceptGraspF};
  \addplot[pBs] table[x=tau, y=Static_sampled, col sep=comma] {\tblecebytauInterceptGraspF};
  \nextgroupplot[xmin=0, xmax=1.32, ymin=0, ymax=0.5, xlabel={launch delay $\tau$ [s]}, title={(d) Block}]
  \addplot[pTIflat] table[x=tau, y=TI, col sep=comma] {\tblecebytauBlockF};
  \addplot[pBone] table[x=tau, y=Static, col sep=comma] {\tblecebytauBlockF};
  \addplot[cBone!45, thick, dashed, mark=square, mark size=1.2pt] table[x=tau, y=StaticTT, col sep=comma] {\tblecebytauBlockF};
  \addplot[pBp] table[x=tau, y=Static_predicted, col sep=comma] {\tblecebytauBlockF};
  \addplot[pBs] table[x=tau, y=Static_sampled, col sep=comma] {\tblecebytauBlockF};
\end{groupplot}
\end{tikzpicture}

\vspace{2pt}
\ref{calleg}
\caption{Calibration of skill-success predictors on held-out launches. (a, b) Reliability on delayed launches
(equal-mass bins). The static model is severely over-confident, a refitted temperature only flattens it, and the
time-indexed model is close to the diagonal. (c, d) ECE against the launch delay. A single temperature improves
calibration at long delays at the cost of calibration at zero delay, and forecast-based corrections reduce the error
without removing it. The sampled-forecast predictor appears in (c, d) only.}
\label{fig:calib}
\end{figure}
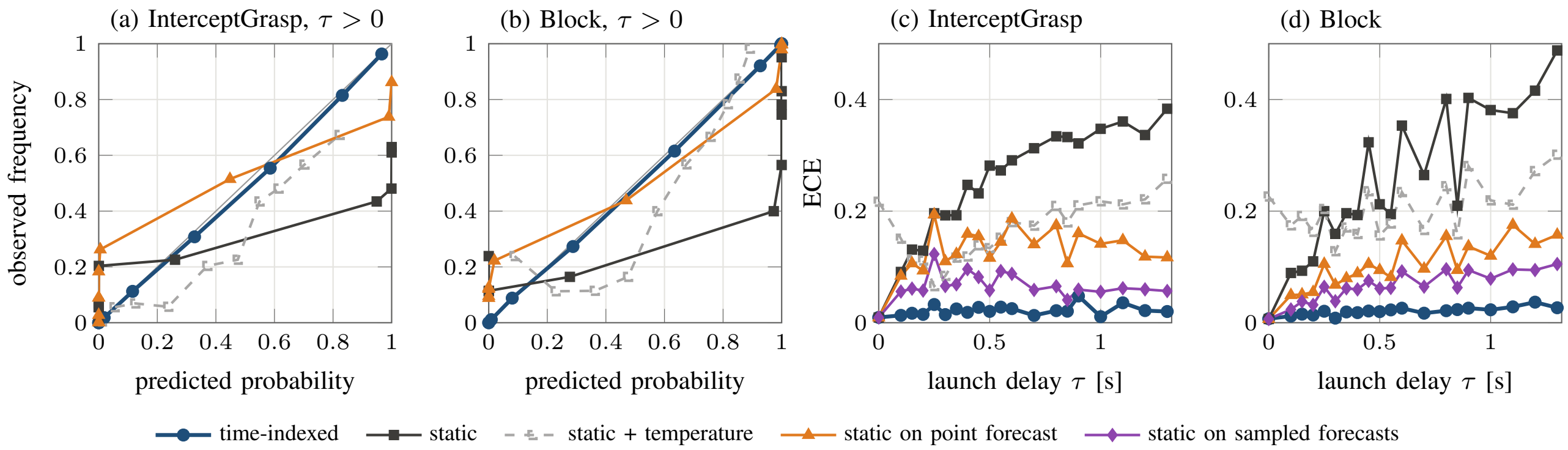
\begin{table}[t]
\centering
\caption{Full calibration metrics on delayed launches (report half of the held-out episodes): ECE with equal-width
and equal-mass bins, Brier score with Murphy's reliability and resolution terms, negative log-likelihood, and mean
prediction minus observed rate (points).}
\label{tab:calibfull}
\setlength{\tabcolsep}{4pt}
\begin{tabular}{@{}llccccccc@{}}
\toprule
Skill, target & Predictor & ECE & ECE (mass) & Brier & Reliability & Resolution & NLL & Over-conf. \\
\midrule
InterceptGrasp, F & Time-indexed & 0.009 & 0.009 & 0.075 & 0.000 & 0.125 & 0.235 & $+$0.7 \\
InterceptGrasp, F & Static & 0.239 & 0.234 & 0.241 & 0.089 & 0.047 & 2.194 & $+$14.4 \\
InterceptGrasp, F & Static + temperature & 0.119 & 0.120 & 0.167 & 0.019 & 0.053 & 0.494 & $+$11.9 \\
InterceptGrasp, F & Static, point forecast & 0.126 & 0.119 & 0.135 & 0.019 & 0.084 & 1.069 & $-$3.3 \\
InterceptGrasp, F & Static, sampled forecasts & 0.059 & 0.057 & 0.113 & 0.004 & 0.091 & 0.514 & $-$3.6 \\
\midrule
InterceptGrasp, G & Time-indexed & 0.008 & 0.008 & 0.063 & 0.000 & 0.098 & 0.196 & $+$0.5 \\
InterceptGrasp, G & Static & 0.197 & 0.197 & 0.197 & 0.078 & 0.041 & 1.282 & $+$14.0 \\
InterceptGrasp, G & Static + temperature & 0.142 & 0.143 & 0.149 & 0.032 & 0.044 & 0.444 & $+$13.6 \\
InterceptGrasp, G & Static, point forecast & 0.082 & 0.077 & 0.101 & 0.008 & 0.068 & 0.645 & $-$3.4 \\
InterceptGrasp, G & Static, sampled forecasts & 0.051 & 0.048 & 0.091 & 0.003 & 0.073 & 0.390 & $-$3.7 \\
\midrule
Block, F & Time-indexed & 0.006 & 0.005 & 0.059 & 0.000 & 0.191 & 0.186 & $+$0.3 \\
Block, F & Static & 0.234 & 0.234 & 0.235 & 0.063 & 0.078 & 2.443 & $+$13.5 \\
Block, F & Static + temperature & 0.133 & 0.134 & 0.178 & 0.027 & 0.099 & 0.544 & $+$8.2 \\
Block, F & Static, point forecast & 0.092 & 0.090 & 0.107 & 0.011 & 0.153 & 0.867 & $-$4.3 \\
Block, F & Static, sampled forecasts & 0.053 & 0.054 & 0.097 & 0.005 & 0.157 & 0.486 & $-$4.6 \\
\midrule
Block, G & Time-indexed & 0.006 & 0.004 & 0.063 & 0.000 & 0.133 & 0.197 & $+$0.1 \\
Block, G & Static & 0.116 & 0.116 & 0.144 & 0.025 & 0.077 & 0.608 & $+$8.2 \\
Block, G & Static + temperature & 0.084 & 0.085 & 0.133 & 0.014 & 0.076 & 0.434 & $+$6.6 \\
Block, G & Static, point forecast & 0.057 & 0.055 & 0.088 & 0.005 & 0.112 & 0.440 & $-$0.3 \\
Block, G & Static, sampled forecasts & 0.028 & 0.028 & 0.081 & 0.001 & 0.116 & 0.310 & $-$1.8 \\
\bottomrule
\end{tabular}
\end{table}

\section{Stale launches}
\label{app:stale}
\begin{table}[t]
\centering
\caption{Stale InterceptGrasp launches (the skill finds no feasible intercept when it starts) as a share of
episodes (\%), and the avoidable subset in which WaitPick or Block launched at the same delay would have completed
the task.}
\label{tab:stale}
\setlength{\tabcolsep}{4pt}
\begin{tabular}{@{}lcccccccc@{}}
\toprule
 & \multicolumn{4}{c}{All stale, $\ell$ [ms]} & \multicolumn{4}{c}{Avoidable stale, $\ell$ [ms]} \\
\cmidrule(lr){2-5}\cmidrule(l){6-9}
Planner & 100 & 150 & 200 & 300 & 100 & 150 & 200 & 300 \\
\midrule
B0 feasibility rule & 5.4 & 12.2 & 12.0 & 14.2 & 3.0 & 7.3 & 7.3 & 7.9 \\
B1 static model & 9.5 & 15.3 & 15.2 & 17.2 & 1.7 & 5.9 & 6.2 & 6.3 \\
B1p static, point forecast & 7.0 & 10.0 & 9.2 & 6.7 & 1.6 & 3.9 & 4.0 & 2.7 \\
B1s static, sampled forecasts & 8.1 & 10.8 & 9.7 & 7.5 & 1.1 & 2.5 & 2.1 & 1.7 \\
TI-fac (ours) & 8.3 & 10.2 & 10.0 & 8.9 & 0.3 & 1.5 & 1.7 & 0.9 \\
TI-flat (ours) & 14.2 & 18.5 & 19.4 & 22.5 & 0.3 & 1.2 & 1.2 & 0.9 \\
\bottomrule
\end{tabular}
\end{table}

\section{Frozen-clock overstatement}
\label{app:frozen}
\begin{table}[t]
\centering
\caption{Success with the world frozen during latency and computation minus success with the clock running (points),
on the same episodes.}
\label{tab:frozen}
\setlength{\tabcolsep}{4pt}
\begin{tabular}{@{}lcccc@{}}
\toprule
 & \multicolumn{4}{c}{$\ell$ [ms]} \\
\cmidrule(l){2-5}
Planner & 100 & 150 & 200 & 300 \\
\midrule
Always WaitPick & $-$0.2 & $-$0.2 & $+$0.2 & $-$0.1 \\
Always InterceptGrasp & $+$3.7 & $+$7.2 & $+$8.7 & $+$11.1 \\
B0 feasibility rule & $+$8.1 & $+$15.9 & $+$17.5 & $+$22.4 \\
B1 static model & $+$7.4 & $+$13.9 & $+$16.8 & $+$20.9 \\
B1p static, point forecast & $+$8.4 & $+$14.8 & $+$17.5 & $+$20.2 \\
B1s static, sampled forecasts & $+$8.6 & $+$13.6 & $+$16.4 & $+$22.8 \\
TI-fac (ours) & $+$7.1 & $+$10.1 & $+$13.0 & $+$18.1 \\
TI-flat (ours) & $+$6.2 & $+$10.6 & $+$13.1 & $+$17.5 \\
\bottomrule
\end{tabular}
\end{table}

\section{Speed bands}
\label{app:speed}
\begin{table}[t]
\centering
\caption{TI-flat minus B1 and minus B1s by launch-speed band [m/s] (points). Bold entries have 95\% intervals above zero.}
\label{tab:speed}
\setlength{\tabcolsep}{2.6pt}
\begin{tabular}{@{}lccccccc@{}}
\toprule
 & & \multicolumn{3}{c}{vs B1, $\ell$ [ms]} & \multicolumn{3}{c}{vs B1s, $\ell$ [ms]} \\
\cmidrule(lr){3-5}\cmidrule(l){6-8}
Band & $n$ & 150 & 200 & 300 & 150 & 200 & 300 \\
\midrule
slow, 0.30 to 0.50 & 579 & \textbf{$+$7.9} & \textbf{$+$9.0} & \textbf{$+$6.9} & $+$2.2 & \textbf{$+$4.3} & \textbf{$+$3.1} \\
medium, 0.50 to 0.75 & 726 & \textbf{$+$5.0} & \textbf{$+$5.1} & \textbf{$+$6.2} & \textbf{$+$3.6} & \textbf{$+$4.3} & \textbf{$+$8.0} \\
fast, 0.75 to 1.00 & 695 & \textbf{$+$2.7} & \textbf{$+$2.6} & \textbf{$+$2.4} & \textbf{$+$4.2} & \textbf{$+$2.7} & \textbf{$+$5.3} \\
\bottomrule
\end{tabular}
\end{table}

\section{Hardware test}
\label{app:hw}
Table~\ref{tab:hw} summarises the hardware test, Table~\ref{tab:hwparams} lists the set-up, the hardware-matched
simulator and its calibration, and Table~\ref{tab:hwwindow} compares the forecast-based and time-indexed predictors
inside the launch window of the hardware protocol.

\begin{table}[t]
\centering
\caption{Hardware test on 49 held-out launches. The confident delayed subset contains the 35 launches with $\tau=0.2$\,s and a static prediction of at least 0.8 (mean prediction, and observed
rate with its 95\% Wilson interval). $\Delta$Brier is the difference from the time-indexed model, with 95\% intervals
from 5{,}000 bootstrap resamples of launches. $^{\dagger}$Uses the frames recorded during the delay, which are not
available when deciding.}
\label{tab:hw}
\footnotesize
\setlength{\tabcolsep}{2.5pt}
\begin{tabular}{@{}lcccc@{}}
\toprule
Predictor & Conf.\ delayed & Brier & $\Delta$Brier [95\%] & AUC \\
\midrule
Static $F(z)$ & 0.99 & 0.260 & $+$0.230 \scriptsize[0.110, 0.360] & 0.84 \\
$F$, forecast state & 0.60 & 0.020 & $-$0.011 \scriptsize[$-$0.021, $-$0.003] & 1.00 \\
$F$ at launch$^{\dagger}$ & 0.61 & 0.006 & $-$0.024 \scriptsize[$-$0.056, $-$0.005] & 1.00 \\
Time-indexed $\G$ & 0.56 & 0.031 &  & 1.00 \\
\midrule
Observed & 0.63 \scriptsize[0.46, 0.77] & & & \\
\bottomrule
\end{tabular}
\end{table}

\begin{table}[h]
\centering
\caption{Hardware test: set-up, hardware-matched simulator, calibration and protocol.}
\label{tab:hwparams}
\small
\begin{tabular}{@{}llp{10.4cm}@{}}
\toprule
Group & Item & Value \\
\midrule
Hardware & robot & UR5e, Cartesian servo streaming at 125\,Hz (look-ahead 0.05\,s, gain 1000) \\
 & tool & Robotiq Hand-E, fingers fully open (53\,mm), bag behind the fingers; lateral clearance 5.2\,mm \\
 & task & golf ball (42.7\,mm) rolling on a 58.4\,cm by 48.7\,cm board inclined by about $3^\circ$, released by hand from rest \\
 & perception & RealSense D455f, 90 frames/s, colour detection, robot-calibrated board frame (about 1\,mm) \\
 & estimation & extended Kalman filter on a fitted acceleration field (25 parameters, fit RMS 2.5\,mm on 72 rolls) \\
 & skill & lateral move along the catch line to the predicted crossing, time-optimal with $v\leq0.5$\,m/s and $a\leq2$\,m/s$^2$, hand between $y=12$ and 44.5\,cm \\
Protocol & decision & when the predicted time to the line falls below a threshold drawn from $U(0.45,0.85)$\,s \\
 & delay & $\tau=0$ (probability $1/4$) or $0.2$\,s ($3/4$); the hand holds its ready pose until launch \\
 & sets & calibration: 35 launches; models frozen; test: 52 launches, of which 3 discarded (control-loop interval $>20$\,ms) and 1 relabelled
(first-contact rule: a ball stopped at the fingers is a miss) \\
Simulator & world & fitted field with Ornstein-Uhlenbeck accelerations ($\sigma$ 0.04\,m/s$^2$ lateral, 0.02 along the slope, 0.8\,s), tracker noise 0.5\,mm and latency 10\,ms, camera-robot bias $\sigma$ 1.5\,mm \\
 & data & 5{,}000 episodes, 14 decisions $\times$ 8 launch delays (0 to 0.4\,s), 449{,}744 exact pairs; 30\% of episodes held out \\
 & models & gradient-boosted classifiers \cite{pedregosa2011scikit}: $F$ on zero-delay pairs, $\G$ on all pairs; ten
features (filtered position and velocity, their uncertainties, predicted crossing time and position, distance to the
ready pose, time since release) \\
Calibration & arm & delay 35\,ms and first-order lag 25\,ms behind the streamed target \\
 & capture & decided when the ball centre is 3.9\,cm before the taught line (median 78\,ms before the predicted
crossing); tolerance 16\,mm with a per-episode spread of 2.4\,mm (largest capture 15.6\,mm, smallest miss 16.9\,mm on the calibration set) \\
 & safety & target frozen once the ball reaches the hand; hand limited to $y\geq12$\,cm (shoulder speed) \\
\bottomrule
\end{tabular}
\end{table}

\begin{table}[h]
\centering
\caption{Forecast against time index inside the launch window of the hardware protocol (launch 0.3 to 0.7\,s before
the line) in the hardware-matched simulator. The last column gives the number of paired launches needed to detect the
difference with a two-sided test at the 5\% level and 80\% power.}
\label{tab:hwwindow}
\small
\begin{tabular}{@{}lcccccc@{}}
\toprule
World & $\tau$ [s] & Launches & Brier, forecast & Brier, time-indexed & Difference & Launches for 80\% power \\
\midrule
nominal & 0.2 & 6{,}119 & 0.0268 & 0.0257 & $+$0.0011 & 14{,}339 \\
nominal & 0.3 & 6{,}154 & 0.0297 & 0.0298 & $-$0.0001 & no gap \\
nominal & 0.4 & 6{,}074 & 0.0380 & 0.0359 & $+$0.0021 & 6{,}407 \\
doubled perturbations & 0.2 & 6{,}083 & 0.0458 & 0.0429 & $+$0.0030 & 4{,}411 \\
doubled perturbations & 0.3 & 6{,}160 & 0.0596 & 0.0554 & $+$0.0042 & 3{,}155 \\
doubled perturbations & 0.4 & 6{,}155 & 0.0771 & 0.0706 & $+$0.0065 & 2{,}043 \\
\bottomrule
\end{tabular}
\end{table}

\end{document}